%% file: main.tex
\documentclass[letterpaper, 10pt, conference]{ieeeconf}      

\IEEEoverridecommandlockouts                              

\usepackage{graphics} 
\usepackage{graphicx}
\usepackage[percent]{overpic}
\usepackage{tikz}
\usepackage{color}
\usepackage{subcaption}
\usepackage{booktabs} 
\usepackage[british]{babel}
\usepackage{multirow}
\usepackage[hyphens]{url}
\usepackage{hyperref}
\usepackage{algorithm}
\usepackage{algpseudocode}
\usepackage{cite}
\usepackage{epsfig} 
\usepackage{amsmath,amssymb}
\usepackage{multirow}
\usepackage[font=small]{caption} 
\usepackage{subcaption}
\usepackage{multirow}
\usepackage{url}
\usepackage{comment}
\usepackage{gensymb}
\usepackage[table]{xcolor}

\graphicspath{ {./images/} }
\DeclareGraphicsExtensions{.png,.pdf,.jpeg,.jpg}

\usepackage{todonotes}
\usepackage{soul}
\definecolor{smoothgreen}{rgb}{0.7,1,0.7}
\sethlcolor{smoothgreen}

\definecolor{darkpink}{rgb}{0.91, 0.33, 0.5}

\title{\LARGE \bf Two-Stage Extrinsic Calibration of a Static Line-Scanning Lidar with a Rotary Platform}

\author{Vikram Shree, Hike Danakian, Long Nguyen, Rajanish Gokidi, Patrick Nercessian
\thanks{The authors are with SiLC Technologies, Monrovia CA, USA. Email: {\small \tt \{vikram, hike, nguyenl, gokidi, patrick\} @silc.com}. The manuscript has been accepted to the 2026 IEEE/RSJ International Conference on Intelligent Robots and Systems (IROS).}
}

\begin{document}

\maketitle


\begin{abstract}
    \label{sec:abstract}
      A line-scanning lidar yields range and azimuth values in a fixed plane. To perceive surrounding objects in 3D, there must be relative motion between the lidar plane and the object. Thus, using a rotating base-platform is promising for industrial applications where objects need to be scanned or inspected precisely, and is the main focus of this work. In the rotary platform setup, a 3D point cloud of an object can be constructed if the axis of rotation and the precise motion about that axis are known. However, this setup gives rise to the following problem: how can the axis of rotation of the platform be accurately identified with respect to the lidar coordinate system? It is referred to as the calibration problem in the robotics community. Any inaccuracy in this transformation directly affects the quality of the reconstructed point cloud, leading to misrepresentation of the object of interest. In this work, we explore automated approaches to statically and dynamically estimate the transformation of a rotary platform's axis of rotation with respect to a static line-scanning lidar. The proposed algorithms have been validated on real-world datasets obtained from a custom made rotary platform and an FMCW lidar, and their convergence characteristics are studied for various initial conditions.
\end{abstract}


\input{contents/introduction}
\input{contents/lit_review_updated}
\input{contents/problem_formulation}
\input{contents/algorithms}

\input{contents/initial_calibration}
\input{contents/finetune}

\input{contents/implementation}

\input{contents/experiments}
\input{contents/conclusion}
\section*{ACKNOWLEDGMENT}
The authors thank Dharma Hariharan Babu and Pete Chayapirad of
SiLC Technologies' mechanical team for their support in building the rotary platform.


\bibliographystyle{IEEEtran} 
\bibliography{egbib} 

\end{document}

%% file: contents/introduction.tex
\section{Introduction}
\label{sec:introduction}

In recent years, the field of robotics has benefited tremendously from advances in 3D perception systems, such as time-of-flight (ToF), stereo vision, structured light, and lidar \cite{o2018computer}. Although there are trade-offs to each of these sensing technologies, lidar has shown promise due to its low latency, superior range precision, low light performance, and wider range of operation spanning from a few microns to several kilometers \cite{behroozpour2017lidar}. Consequently, the past decade has seen a surge in lidar usage for robotics, 3D scanning, machine inspection, defect detection, ADAS applications, topography, geoscience, and atmospheric studies \cite{roriz2021automotive}, \cite{poenicke2023industrial}. 

2D lidar systems commonly employ electromechanical scanning mechanisms to extend the dimensionality of their measurements (e.g., Sweep, Terabee, and Rpilidar \cite{electronics9050741}), enabling deployment across a wide range of indoor and outdoor robotic applications. For object scanning and 3D model reconstruction, however, a common approach is to place the object on a rotating turntable while a static lidar performs planar scans \cite{wang2022calibration}. Generating an accurate 3D point cloud from these 2D scans requires precise knowledge of the coordinate transformation between the lidar frame and the axis of rotation. Consequently, the quality of the reconstructed point cloud depends directly on this transformation. Estimating these extrinsic parameters, commonly referred to as calibration, is the primary focus of this work.


Given its direct impact on point cloud accuracy, the calibration problem has been extensively studied. Prior work has largely focused on the automatic estimation of extrinsic parameters for multi-beam spinning lidar systems \cite{sheehan2012self,levinson2014unsupervised}, motivated by their widespread use in autonomous driving. These approaches typically require no dedicated calibration target and instead exploit the presence of multiple planar surfaces observed during a full rotational sweep of the environment. Similarly, other methods leverage measurement redundancy inherent in a complete lidar rotation to formulate cost functions for pose estimation \cite{alismail2015automatic, claer2019calibration, yamao2017calibration}.

However, these techniques are not directly applicable to the calibration of a static lidar observing a rotary platform. In this setting, the rotating platform presents only a single planar surface with limited geometric features, rendering such approaches ineffective. To address this, some researchers have proposed semi-automatic calibration methods that rely on specific target geometries to estimate the extrinsic parameters of a static lidar with respect to a turntable \cite{schmidt2022doe,zhu2019analytic}. Nevertheless, these methods typically depend on precise operator intervention, making them prone to human error, or assume a predefined orientation between the lidar and the rotary platform \cite{wang2022calibration}, which limits their generalizability to arbitrary sensor configurations.

    
    
    
In this work, we propose an automatic two-stage pipeline for estimating the extrinsic parameters of a static lidar with respect to a rotating platform, without imposing any assumptions on the lidar pose. The first stage performs static calibration to obtain an initial estimate of the transformation parameters. These estimates are subsequently refined through a nonlinear optimization-based dynamic calibration procedure. The key contributions of this paper are:
\begin{enumerate}
    \item A geometry-driven static initial calibration algorithm that computes an estimate of the lidar-to-rotation-axis transformation using target shape constraints.
    \item Two optimization objectives for dynamic refinement: (i) a distortion-minimization cost function that penalizes geometric inconsistencies in the reconstructed 3D point cloud, and (ii) a periodicity-based cost function that exploits measurement redundancy over a full rotation of the platform. We analyze the applicability of each objective to different target geometries and provide a comparative evaluation of their performance.
    \item The design and fabrication of a rotary platform for controlled real-world experimentation, along with extensive Monte Carlo simulations to evaluate robustness under varying lidar poses and initialization conditions. 
\end{enumerate}

%% file: contents/lit_review_updated.tex
\section{Related Work}
\label{sec:related_work}

Calibration between a line-scanning sensor and a rotation axis arises in
two dual configurations: actuated 2D lidars, where the sensor is rotated
to acquire 3D data, and turntable systems, where a static sensor observes
a rotating object. For the former, plane-based extrinsic calibration has
been studied extensively: Kang and Doh \cite{kang2016full} first recovered
the full 6-DOF transformation from a single plane by decoupling the
estimation of rotation from translation, and subsequent formulations
construct separate cost functions for the rotational and translational
parameters \cite{gao2019calibration}. These methods rely on the sensor
sweeping large environmental planes across many rotation angles ---
constraints unavailable in our dual configuration, where the environment
is static in the lidar frame and only the platform, a single plane with
limited geometric features, undergoes motion.

A second family of methods dispenses with dedicated targets by exploiting
measurement redundancy: any structure is observed twice during a full
rotation, and residual misalignment between the redundant observations
exposes calibration error. Yamao et al.~\cite{yamao2017calibration}
minimize this discrepancy for a rotating 2D scanner in unprepared
environments, while Alismail and Browning \cite{alismail2015automatic}
and Claer et al.~\cite{claer2019calibration} formalize the periodicity as
an optimization objective for spinning lidars. Our periodicity-constrained
fine-tuning is directly motivated by these works, with two key
differences: the platform rotates while the lidar is stationary, and the
rotating scene is a featureless disk rather than a structured 3D
environment. We therefore introduce a dedicated target geometry that
restores multi-directional geometric constraints while remaining visible
across both half-scans.

Closest to our physical configuration is the calibration of
turntable-based 3D scanning systems. Camera- and structured-light-based
approaches estimate the turntable axis by registering overlapping views
of a reference tool across rotation increments, typically via ICP-style
alignment \cite{zhu2019analytic}, or tool-free by registering scans of
the natural scene \cite{pang2014tool}. Both fundamentally exploit overlap
between consecutive 2D or 3D fields of view, which does not exist for a
line-scanning lidar that acquires only a one-dimensional profile per
instant. For 2D laser scanners paired with turntables, Wang et
al.~\cite{wang2022calibration} assume a predefined orientation between
the laser plane and the rotation axis, while Schmidt et
al.~\cite{schmidt2022doe} require DOE laser assistance and precise
operator intervention; both assumptions limit applicability to arbitrary
sensor poses. Since the objectives arising in this setting remain highly
nonlinear and initialization-sensitive, we adopt a two-stage strategy: a
geometry-driven static initialization that places the estimate within the
basin of attraction of the global optimum, followed by dynamic nonlinear
refinement based on shape or periodicity constraints.

%% file: contents/problem_formulation.tex
\section{Problem Formulation}
\label{sec:problem_formulation}

\begin{figure}[t]
  \centering
  \includegraphics[width=0.65\linewidth]{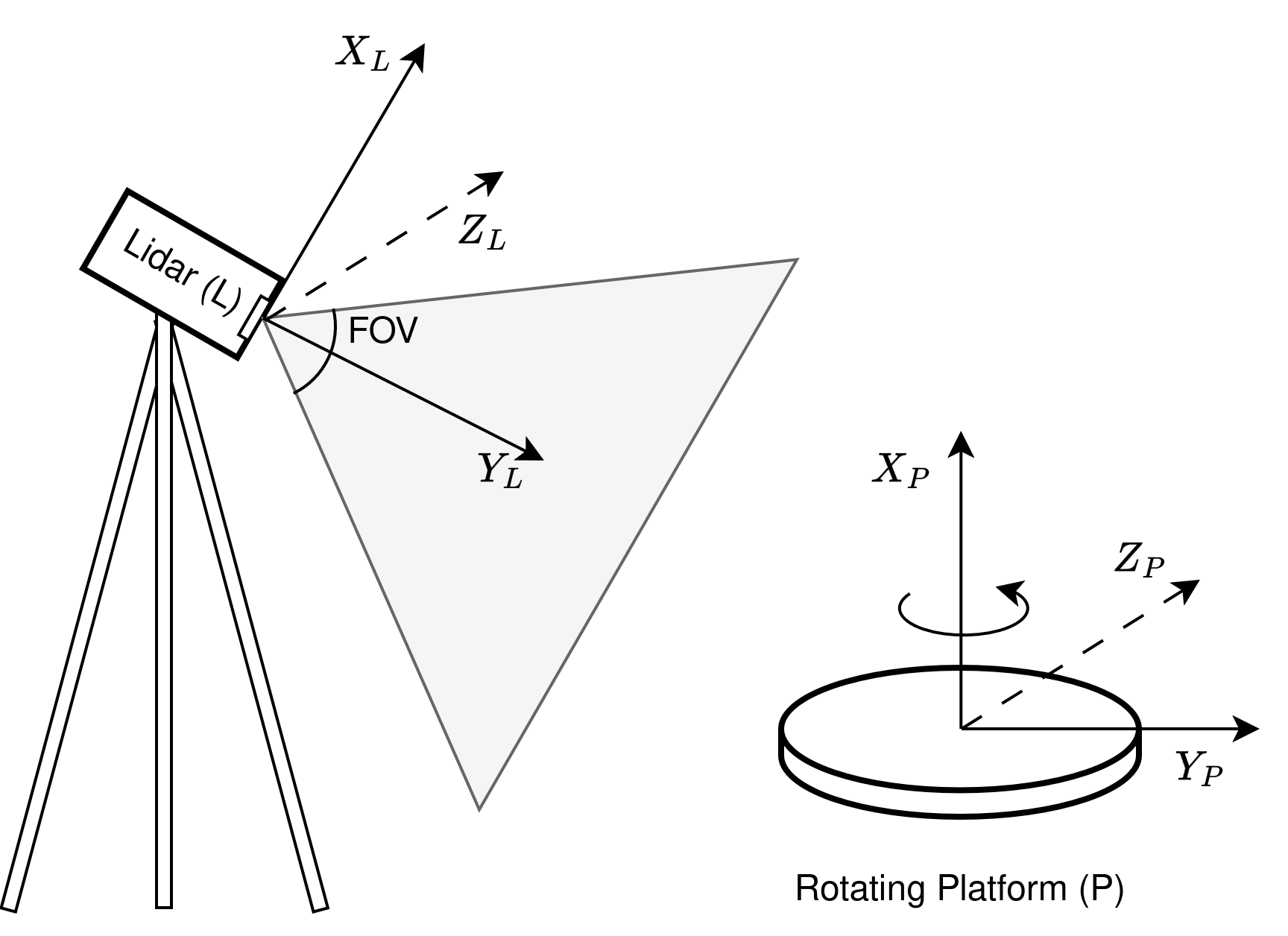}
  \caption{Lidar and rotary platform coordinate system. The $z$-axis for both: lidar, and the platform, is into the plane of the paper. }
  \label{fig:trace_rot_axes}
  \vspace{-0.2in}
\end{figure}

Fig.~\ref{fig:trace_rot_axes} shows two axes systems: the line-scanner ($L$) and the rotating platform ($P$), where the platform is assumed to be rotating about the $X_{P}$ axis. Both are static in the global frame of reference. The calibration problem can be represented as a 6-DOF estimation problem, where the variables to be identified are translation of frame-$P$ $\{ t_x, t_y, t_z\}$ and the orientation of frame-$P$, represented by Euler angles $\{r_x, r_y, r_z\}$, with respect to frame-$L$. These parameters are denoted by $\mathbf{H}$. Since the disk is free to rotate along its axis, we can freely choose $r_{x}$ without any loss of generality, reducing its degrees-of-freedom to 5. 

%% file: contents/algorithms.tex
\section{Algorithms}
\label{sec:algorithms}


The estimation of calibration parameters constitutes a nonlinear optimization problem. To address this, we adopt a two-stage calibration strategy. The first stage performs a coarse initialization to obtain an approximate estimate of the extrinsic parameters $\{t_x, t_y, t_z, r_y, r_z\}$. These initial estimates are subsequently refined in a second stage through nonlinear optimization to minimize a cost function. The details of both stages are presented in the following sections.


%% file: contents/initial_calibration.tex
\subsection{Initial Calibration}
\label{subsec:initial_calibration}
To simplify the initial calibration problem, we make the following assumption. 

\textit{Assumption} 1: The roll angle of frame-$P$ w.r.t frame-$L$ is negligible, i.e. $r_y \approx 0$.

\begin{figure}[t]
  \centering
  \begin{subfigure}[b]{0.23\textwidth}
    \centering
    \includegraphics[width=0.8\linewidth, trim={0cm 0cm 0cm 0cm}, clip]{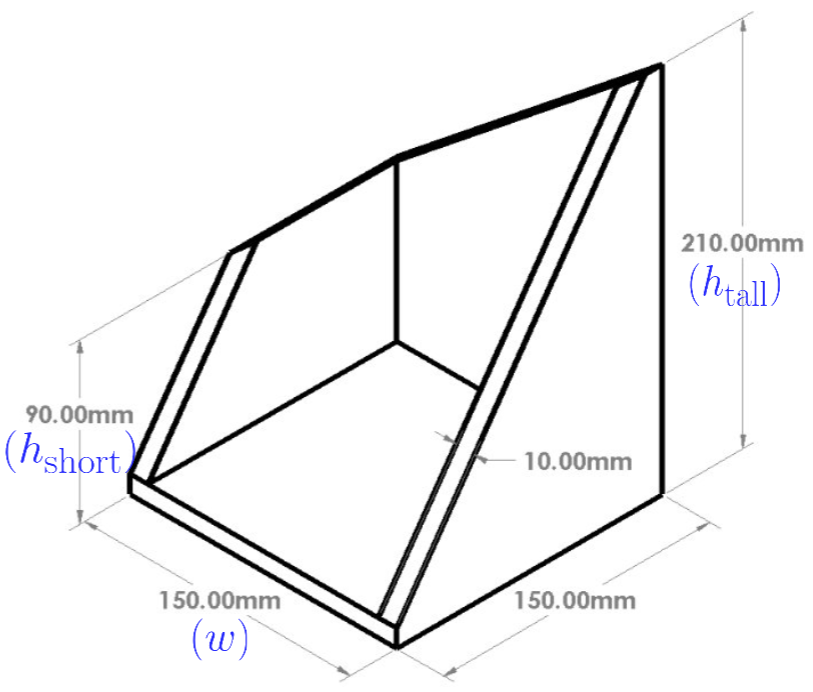}
    \caption{CAD model.}
    \label{fig:initial_calib_cad}
  \end{subfigure}
  \hfill
  \begin{subfigure}[b]{0.23\textwidth}
    \centering
    \includegraphics[width=0.55\linewidth, trim={1.5cm 0cm 1.5cm 0cm}, clip]{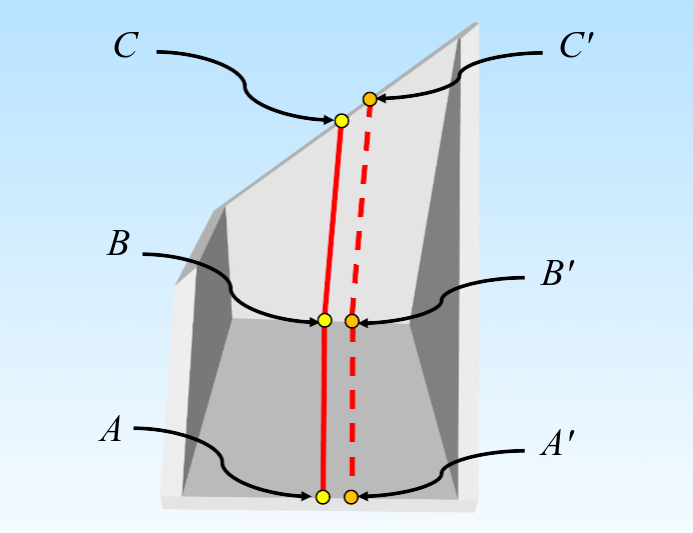}
    \caption{Target orientation and lidar's plane of incidence.}
    \label{fig:initial_calib_cad_labelled}
  \end{subfigure}
  \caption{Initial calibration target. $A$, $B$, and $C$ denote the center of the respective edges of the target, and constitute a hypothetical plane of  incidence when lidar's scanning plane intersects the rotary axis. The dashed line denotes the actual intersection of lidar's plane of incidence with the edges of the target. $B$ must align with the center of the rotary platform.}
  \label{fig:initial_calib_target}
  \vspace{-0.2in}
\end{figure}

\textit{Assumption} 1 allows us to use a simple geometric shape to assess the remaining calibration parameters $\{t_x, t_y, t_z, r_z\}$.
Fig.~\ref{fig:initial_calib_cad} shows a CAD model for the shape of the calibration target that we use for calibration. During initial calibration, the target is kept statically on top of the rotary platform such that it faces the lidar and its midpoint $B$, depicted in Fig.~\ref{fig:initial_calib_cad_labelled}, aligns with the center of the platform. The lidar's scanning plane intersects the target's edges, say at $A'$, $B'$, and $C'$. Identifying these points in the line-scan data allows us to estimate the extrinsic parameters. 

An overview of the algorithm is shown in Algorithm~\ref{alg:initial_calibration_algo}. It involves two key steps: azimuth-bin smoothing and contiguous line detection. The line-scanner data stream is a sequence of range-azimuth data, denoted by $\{\mathbf{p}_i\} \equiv \{\{\rho_i, \alpha_i\}\}$. In the former step, the line-scan data is converted into discrete azimuth bins with a median filter to reject outliers. In the latter step, line detection is used to get a set of consecutive lines $\{\mathbf{l}_k\}$. The geometry of the CAD model is used to identify candidate lines on the target $\{\mathbf{l}_{A'B'}, \mathbf{l}_{B'C'}\}$, allowing us to estimate the parameters $\{t_x, t_y, t_z, r_z\}$.

\subsubsection{Estimating $r_z$, $t_x$, and $t_y$}
Given the lidar data on the target shown in Fig.~\ref{fig:initial_calib_cad_labelled}, the pitch angle $r_z$ can be computed based on the geometry of line $\mathbf{l}_{A'B'}$. The polar coordinates of points $A'$ and $B'$ are denoted as $\mathbf{p}_{A'}(\rho_{A'}, \alpha_{A'})$ and $\mathbf{p}_{B'}(\rho_{B'}, \alpha_{B'})$, respectively. Then,
\begin{equation}
    r_z = -\arctan\bigg(\frac{\rho_{B'} \sin\alpha_{B'} - \rho_{A'} \sin\alpha_{A'}}{\rho_{B'} \cos \alpha_{B'} - \rho_{A'}\cos \alpha_{A'}}\bigg).
    \label{eq:pitch_angle}
\end{equation}
The center of the platform is equivalent to the coordinates of $B'$ in lidar's frame of reference $L$ i.e.
\begin{equation}
    t_x = \rho_{B'}\sin\alpha_{B'}, \ \text{and} \  t_y = \rho_{B'}\cos\alpha_{B'}.
    \label{eq:center_coord}
\end{equation}
For readability, the CAD model thickness is omitted from the equation but must be accounted for when determining the translation and rotation parameters.

\subsubsection{Estimating $t_z$}
The asymmetric back wall of the target allows estimating translation parameter $t_z$ based on the observed height of $\mathbf{l}_{B'C'}$. Assuming that $h_{\text{short}}, h_{\text{mid}}, h_{\text{tall}}$ represent the vertical short-edge, mid-edge and tall-edge, of the target, and $w$ denotes the width of the target, we get,
\begin{equation}
    t_z = \frac{w(|\mathbf{l}_{B'C'}| - h_{\text{mid}})}{h_{\text{tall}} - h_{\text{short}}}
    \label{eq:z_offset}
\end{equation}
Thus, a longer edge $|\mathbf{l}_{B'C'}|$ compared to $h_{\text{mid}}$ implies a positive value for $t_z$, and vice versa. Note that \textit{Assumption} 1 may introduce errors in estimating the parameters; however, the goal of initial calibration is to primarily provide a reliable initial guess for the fine-tuning algorithm, which would then further refine the estimate.


\begin{algorithm}[htbp]
\caption{Overview of initial calibration algorithm}
\label{alg:initial_calibration_algo}
\begin{algorithmic}[1]  
  \Require Raw lidar data $\{\mathbf{p}_i\}$ and target dimensions
  \State $\{\mathbf{p}'_{j}\}$ = AzimuthBinSmoothing($\{\mathbf{p}_i\}$)
  \State $\{\mathbf{l}_k\}$ = FindContigousLines($\{\mathbf{p}'_j\}$)
  \State $\{\mathbf{l}_{A'B'}, \mathbf{l}_{B'C'}\}$ = IdentifyCandidateLines($\{\mathbf{l}_{k}\}$)
  \State $\{t_x, t_y, t_z, r_z\}$ = EstimateExtrinsics($\{\mathbf{l}_{A'B'}, \mathbf{l}_{B'C'}\}$)
\end{algorithmic}
  \Return $\{t_x, t_y, t_z, r_z\}$
\end{algorithm}

%% file: contents/finetune.tex
\subsection{Fine-tuning Calibration}
\label{subsec:finetune}
The goal is to further refine the estimates for the calibration parameters that the initial calibration algorithm yields. 
The initial calibration algorithm, due to its reliance on a static target placement, is prone to errors. For example, the target may not be placed properly on the rotary platform during the process or there can be minor misalignment between the platform's vertical axis and the actual rotational axis. To compensate for these sources of error, we explore two dynamic fine-tuning algorithms, both based on the principle of cost function minimization.

\subsubsection{Shape-constrained fine-tuning}
\label{subsubsec:shape_constrained}
\begin{figure}[t]
  \centering
  \begin{subfigure}[b]{0.45\linewidth}
    \centering
    \includegraphics[width=0.6\linewidth]{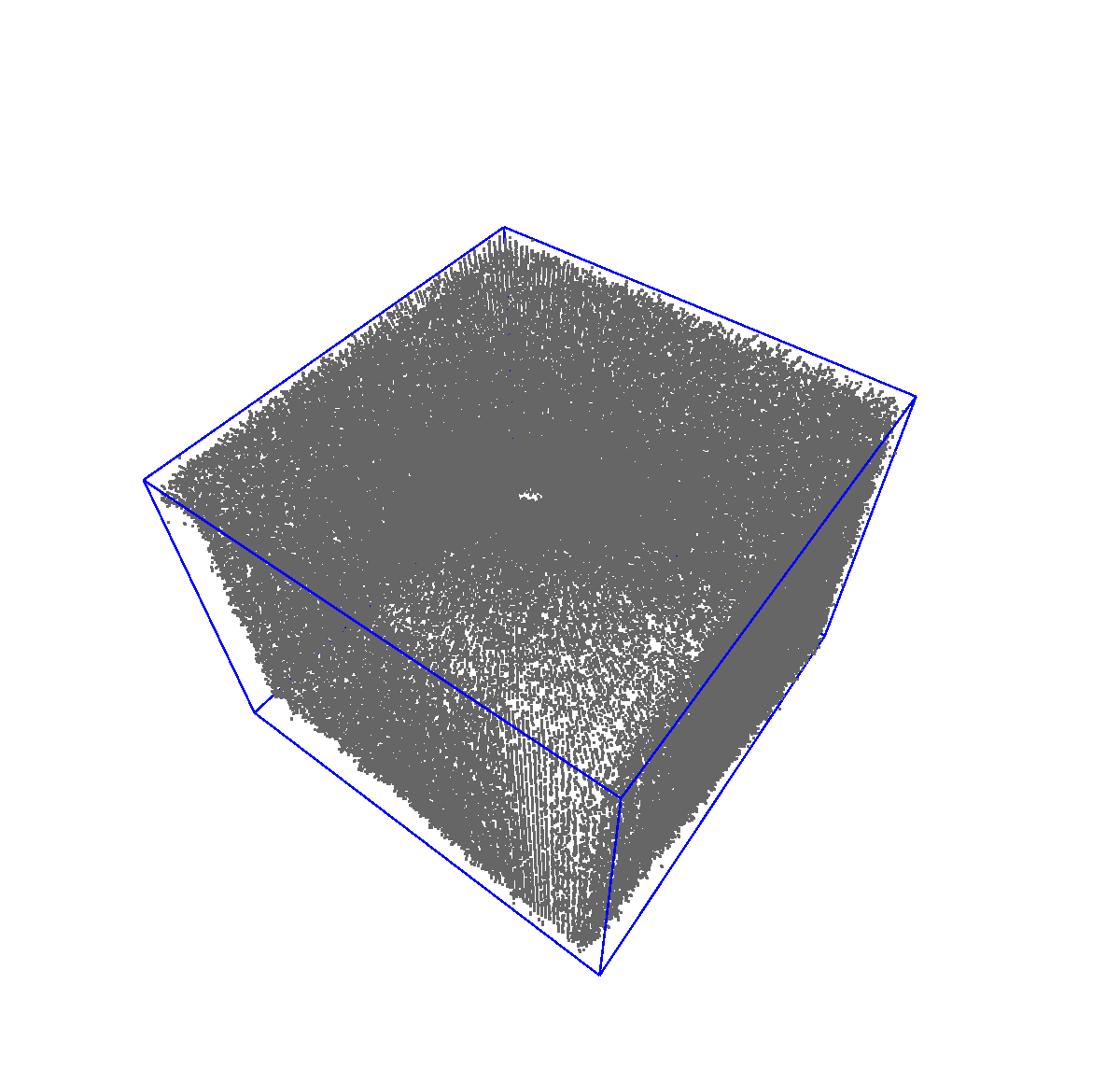}
    \caption{Point cloud obtained from the best estimate of parameters $\mathbf{H^{*}}$.}
    \label{fig:regumal_box_pc}
  \end{subfigure}
  \hfill
  \begin{subfigure}[b]{0.45\linewidth}
    \centering
    \includegraphics[width=0.6\linewidth]{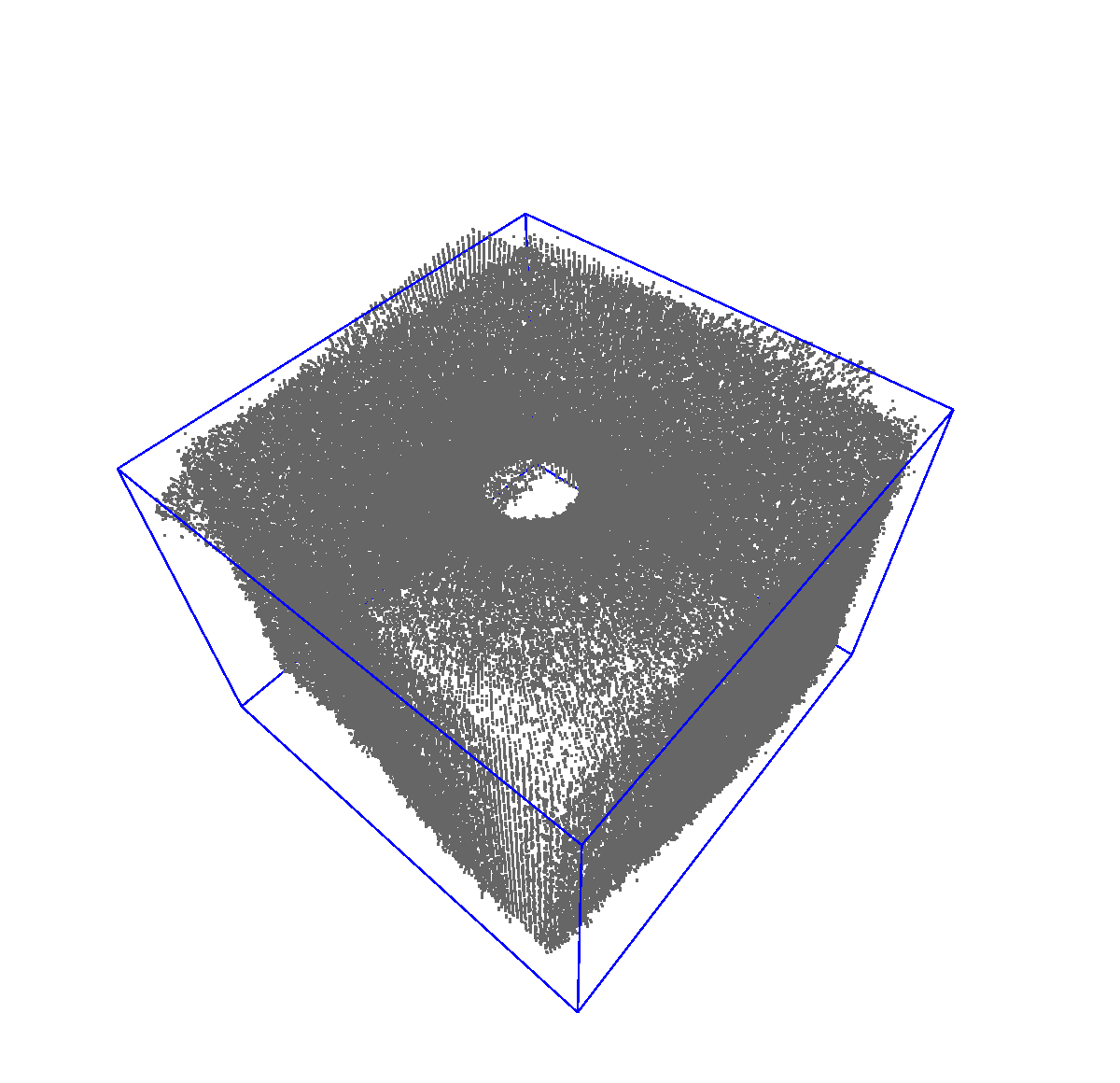}
    \caption{Point cloud obtained from a slightly perturbed parameters $\mathbf{H'}$.}
    \label{fig:distorted_box_pc}
  \end{subfigure}
  \caption{Different point clouds obtained for a 10 cm cube, in a rotary platform setup, with different extrinsic calibration parameters.}
  \label{fig:calibration_distortion_pc}
\end{figure}

On a dynamic rotary platform, the point cloud reconstruction for an object is a function of the choice of calibration parameters $\mathbf{H}$. Fig.~\ref{fig:calibration_distortion_pc} shows point clouds for a cube, given two different calibration parameters. We find that as we approach the \textit{true} calibration parameters, the shape and dimension of the point cloud starts to resemble a cube. In the shape-constrained approach, we leverage the deviation of the point cloud from its \textit{true} shape to quantify the \textit{error} in the calibration parameters. The goal is to create a cost function whose optimization would yield an estimate of the calibration parameters $\mathbf{H}$.


Starting with an approximate $\mathbf{H}$, we can obtain a point cloud for a cube of edge length $s$. To quantify the deviation from its cubic shape, we propose a composite cost function consisting of two terms $C_1$ and $C_2$. While $C_1$ penalizes for deviation in the point cloud from the standard cubic shape, on the other hand, $C_2$ penalizes any deviation from the \textit{true} dimensions of the cube. To create this cost function, we first fit a minimum volume cuboid to the point cloud. A face of the fitted cuboid is denoted by $(\mathbf{c}_i, \hat{\mathbf{n}}_i)$, where $\mathbf{c}_i$ is the coordinate of the center of the face and $\hat{\mathbf{n}}_i$ denotes the inward unit-normal to that face. Given the point cloud set $\{\mathbf{x}_k\}$, we formulate the following cost function: 
\begin{equation}
    \begin{aligned}
    \mathbf{H}^{*} &= \arg\min_{\mathbf{H}} \big(C_1 + C_2\big), \quad \text{where} \\
    C_1 &= \frac{1}{N} \sum_{k=1}^{N} g(\mathbf{x}_k, \mathbf{c}_1, \hat{\mathbf{n}}_1, \cdots; \mathbf{H}), \quad
    C_2 = \sum_{i=1}^{3} \lVert s_i - s \rVert^2 .
    \end{aligned}    
    \label{eq:shape_constrain}
\end{equation}


The first term in Eq.~(\ref{eq:shape_constrain}), denoted by function $g(\cdot)$, identifies the signed distance of a point in the point cloud from the closest face of the cuboid. The second term measures the deviation of edge length from the \textit{true} size of the cube. 


Algorithm~\ref{alg:shape_constrained_cost} outlines the procedure for constructing the shape-constrained cost function. Given the calibration parameters $\mathbf{H}$, the raw lidar points are first transformed into the rotary platform frame, as symbolized by TransformLToP(). A threshold-based segmentation step then classifies the points into three categories: cube, disk, and outliers. An oriented bounding box is subsequently fitted to the points identified as belonging to the cube, and the cost function is formulated according to Eq.~(\ref{eq:shape_constrain}).

The inclusion of segmentation, bounding-box fitting, and signed-distance computations renders the resulting objective highly nonlinear and non-convex. Consequently, parameter refinement is performed using nonlinear least-squares optimization methods, such as the Trust-Region Reflective algorithm or the Levenberg–Marquardt algorithm. These iterative methods minimize the objective by locally approximating the cost landscape and are therefore sensitive to initialization.


\begin{algorithm}[htbp]
\caption{Overview of cost function creation for shape-constrained fine-tuning algorithm}
\label{alg:shape_constrained_cost}
\begin{algorithmic}[1]  
  \Require Raw lidar data $\{\mathbf{p}_{i}\}$, encoder data $\{\phi_{i}\}$, cube dimension $s$, and calibration parameters $\mathbf{H}$
    \State $\{\mathbf{x}_i\}$ = TransformLToP($\{\mathbf{p}_{i}\}$, $\{\phi_{i}\}$, $\mathbf{H}^{(0)}$)
    \State $\{\mathcal{X}^{\text{disk}}, \mathcal{X}^{\text{cube}}, \mathcal{X}^{\text{outliers}}\}$ = SegmentCube($\{\mathbf{x}_i\}$)
    \State $\{(\mathbf{c}_i, \hat{\mathbf{n}}_i)\}, \{s_1, s_2, s_3\}$ = OrientedBoundingBox($\mathcal{X}^{\text{cube}}$)
    \State $\{d_k\}$ = SignedDistFromFace($\mathcal{X}^{\text{cube}}, \{(\mathbf{c}_i, \hat{\mathbf{n}}_i)\}$)
    \State $C_1$ = $\frac{1}{|\{d_k\}|}\sum_{k} d_k^2$
    \State $C_2$ = $\sum_{j=1}^{3} ||s_j-s||^2$
\end{algorithmic}
\Return $C_1 + C_2$
\end{algorithm}

\subsubsection{Periodicity-constrained fine-tuning}
\label{subsubsec:periodicity_constrained}
\begin{figure}[t]
    \centering
    \includegraphics[width=0.9\linewidth]{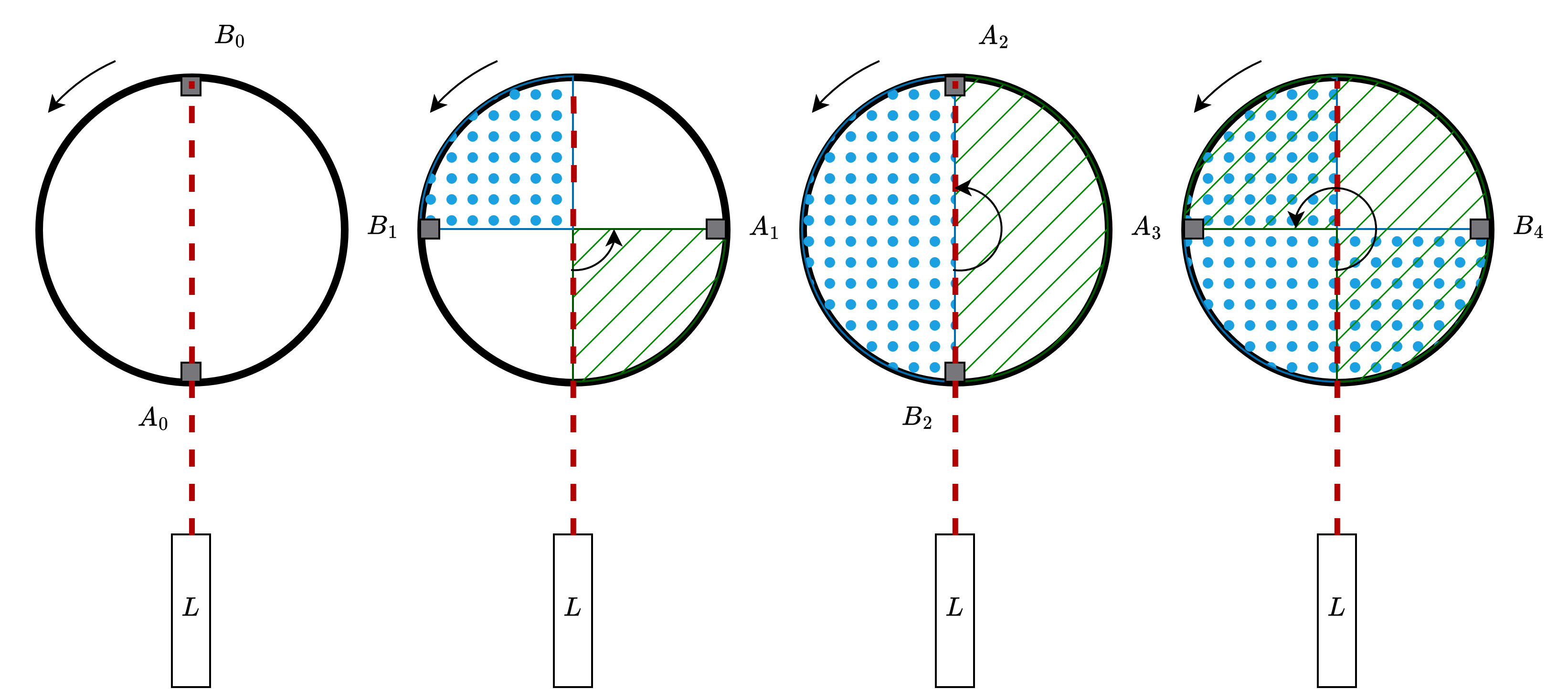}
    \caption{Top-view of the lidar and rotary platform at different configurations, starting with $0\degree$, $90\degree$, $180\degree$, and $270\degree$. The \textcolor{cyan}{dots} depict the portion of the platform being scanned by upper section of the lidar beam, and \textcolor{green}{slanted lines} depict the portion of the plate scanned by the lower portion of the beam. They start to overlap after $180\degree$ rotation of the platform.}
    \label{fig:periodicity_constraint}
\end{figure}

The periodicity in the data was first used in \cite{alismail2015automatic} for calibrating a spinning lidar system. 
In contrast, we have a spinning platform, while the lidar is stationary. Nonetheless, the periodicity argument stays valid and any point on the rotary platform crosses the lidar's plane of incidence twice when it completes one full rotation. Fig.~\ref{fig:periodicity_constraint} shows the regions of the platform being scanned during different rotational configurations and highlights that they start to overlap after half-rotation of the platform. Motivated by the ideas in \cite{alismail2015automatic}, we exploit this periodicity for refining the calibration parameters $\mathbf{H}$. Given an encoder angle $\phi$, we have a set of raw registered points $\mathcal{X}(\phi)$ from the lidar. A full rotation of the platform is divided into two half-scans i.e.
\begin{equation}
     \mathcal{X} = \{ \mathcal{X}(\phi) | 0\leq\phi<\pi\}, \text{and}
\end{equation}
\begin{equation}
     \mathcal{X}' = \{ \mathcal{X}(\phi) | \pi\leq\phi<2\pi\}.
\end{equation}

A physical location on the platform should have a corresponding point in each set $\mathcal{X}$ and $\mathcal{X}'$. 
This periodicity property forms the basis of our objective function, which quantifies the geometric discrepancy between the two sets. Accordingly, the calibration parameters are estimated by solving
\begin{equation}
    \mathbf{H}^{*} = \arg\min_{\mathbf{H}} f(\mathcal{X}, \mathcal{X}'; \mathbf{H}),
    \label{eq:periodicity_cost}
\end{equation}
where $f(\cdot)$ measures the dissimilarity between the corresponding points in the two half-scans. 

Algorithm~\ref{alg:periodicity_algo} provides an explicit formulation of this objective. The procedure consists of four principal steps: (i) transforming raw points from the static lidar frame-$L$ to the static platform frame-$P$, (ii) propagating the points from frame-$P$ to the dynamic platform frame-$D$ ($\{\mathcal{X}_D, \mathcal{X}_D'\}$), (iii) establishing correspondences between $\mathcal{X}_D$ and $\mathcal{X}_D'$, and (iv) estimating surface normals for the matched points. The dissimilarity metric is defined using the point-to-plane distance, which has been shown to provide superior performance in 3D reconstruction and registration tasks \cite{segal2009generalized}.


Unlike spinning lidar systems, which observe rich environmental structure during rotation, a rotary platform without a dedicated calibration target presents only a single planar surface. This configuration does not provide sufficient geometric constraints to solve the 5-DOF calibration problem. 

\begin{figure}[t]
  \centering
  \includegraphics[width=0.5\linewidth,trim={3cm 9cm 2.5cm 3cm}, clip]{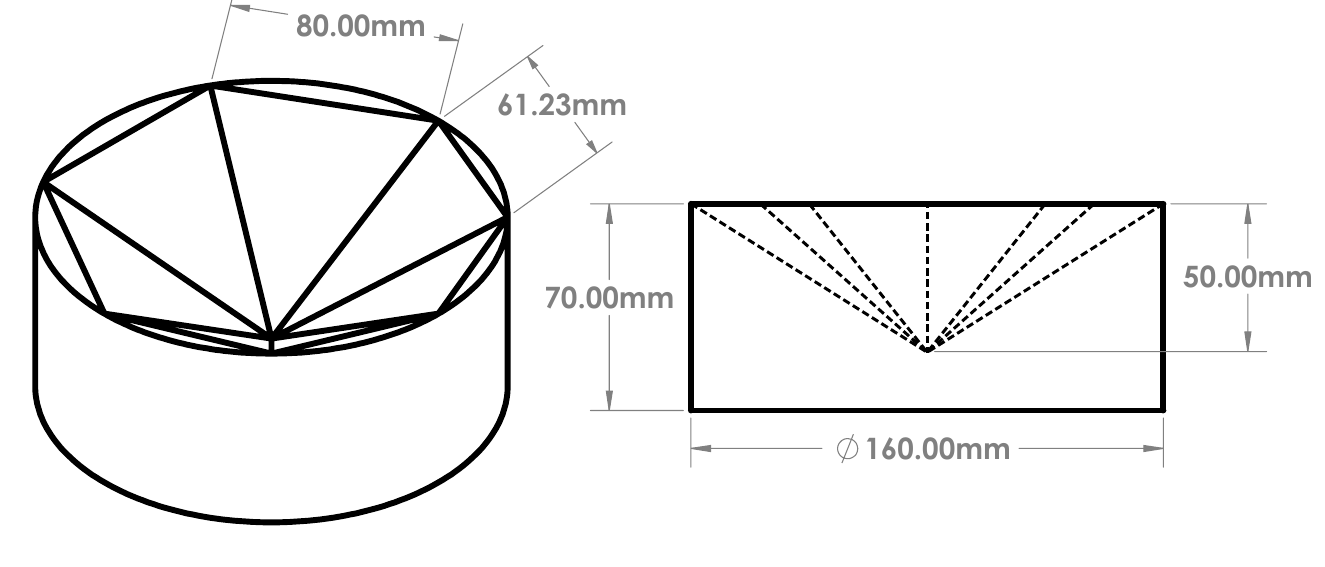}  
  \caption{Heptagonal target used in periodicity-constrained fine-tuning.}
  \label{fig:hep_calib_target}
\end{figure}

To address this limitation, we design a calibration target that introduces geometric variation along multiple directions while ensuring consistent visibility by the lidar across both half-scans. The latter requirement is particularly important during the design phase, as arbitrary shapes (e.g. a cube) offer limited shared surface visibility, typically only the top face, between the two half-scans, thereby weakening correspondence constraints. Accordingly, we employ a concave heptagonal target, shown in Fig.~\ref{fig:hep_calib_target}, which is mounted on the rotary platform during calibration. It is important to note that this geometry is not unique; alternative target shapes may be used, provided they satisfy the criteria of multi-directional geometric constraints and mutual visibility across half-scans.
\begin{algorithm}[ht]
\caption{Overview of periodicity-constrained calibration fine-tuning algorithm}
\label{alg:periodicity_algo}
\begin{algorithmic}[1]  
  \Require $\mathcal{X}$, $\mathcal{X}'$, and $\mathbf{H}^{(0)}$
  \While{not converged}
    \State $\{\mathcal{X}_D, \mathcal{X}_D'\}$ = ApplyCalibration($\mathcal{X}, \mathcal{X}', \mathbf{H}^{(k)}$)
    \State $\{\mathbf{x}_i, \mathbf{x}'_i\}$ = FindNeighbors($\mathcal{X}_D, \mathcal{X}_D';\mathbf{H}^{(k)}$)
    \State $\{\mathbf{\hat{n}}_i, wi\}$ = ComputeWeightedNormals($\mathcal{X}_D;\mathbf{H}^{(k)}$)
    \State $\mathbf{H}^{(k+1)} = \arg\min_{\mathbf{H}} \{ \sum_{k}w_i||\mathbf{\hat{n}}_i . (\mathbf{x}_i - \mathbf{x}'_i)||^{2} \} $
  \EndWhile
\end{algorithmic}
\end{algorithm}

%% file: contents/implementation.tex
\section{Implementation Details}
\label{sec:implementation}
In this section, we detail the key components of the calibration algorithms.

\subsection{Initial Calibration}

During initial calibration, we smooth the data across multiple lidar scans falling onto the target, typically 100, to ensure that outliers are removed from the scan. The AzimuthBinSmoothing() function in Algorithm~\ref{alg:initial_calibration_algo} divides the azimuth into fixed $0.02$ degree bins and the median range is evaluated for each bin. We use Principal Component Analysis (PCA) to identify the lines in the smoothed scan data in the function FindContigousLines(). IdentifyCandidateLines() leverages calibration target dimensions to filter contiguous line segments that intersect at a $90\degree$ angle and satisfy dimensions of the CAD model. Finally, EstimateExtrinsics() relies on Eqs.~(\ref{eq:pitch_angle}), (\ref{eq:center_coord}), and (\ref{eq:z_offset}) to deliver an estimate for the translation parameters and Euler rotation angle. 

\subsection{Shape-constrained fine-tuning}

In Algorithm~\ref{alg:shape_constrained_cost}, to isolate the subset of points corresponding to the cube, denoted by $\mathcal{X}^{\text{cube}}$, we apply a threshold-based segmentation along the $x$-axis of the rotary platform frame-$P$. To reduce misclassification of platform points as cube points, a conservative margin of 1 cm is introduced. While this strategy may exclude a small portion of valid cube points, it ensures that the rotating platform surface is not inadvertently incorporated into $\mathcal{X}^{\text{cube}}$. More advanced segmentation strategies, such as normal-based filtering, could alternatively be employed to remove the rotating disk with higher geometric fidelity.

Outliers are rejected using a radius-based filtering criterion, whereby points with fewer than eight neighbors within a 1 mm radius are discarded. A minimum-volume oriented bounding box is then fitted to $\mathcal{X}^{\text{cube}}$ using the implementation provided in Open3D. For nonlinear optimization, we utilize SciPy’s Trust-Region Reflective algorithm. The search space is constrained to $\pm 2$ cm for translation and $\pm 2^\circ$ for rotational degrees-of-freedom around the initial calibration estimate. The termination tolerances are set to ${\texttt{xTol} = 10^{-12}, \ \texttt{fTol} = 10^{-10}}$, with a maximum of 100 function evaluations.



\subsection{Periodicity-constrained fine-tuning}

In the periodicity-constrained fine-tuning, we use a KD-tree data structure to speed up computation of nearest neighbors with a maximum distance threshold of $20$ cm. To ensure robustness, we enforce a bijective correspondence i.e. $\mathbf{x}_i$ and $\mathbf{x}'_i$ are mutual neighbors of each other. In ComputeWeightedNormals(), the nearest 50 points are chosen to compute the normal direction and the weight is based upon the eigenvalues of the covariance matrix, i.e. $w_i = \frac{2(\lambda_2 - \lambda_3)}{\lambda_1 + \lambda_2 + \lambda_3}$, where $\lambda_1\geq\lambda_2\geq\lambda_3$. Consequently, $w_i\rightarrow1.0$ for planar surfaces. Since the points lying on the rotary platform have only a minor contribution to the optimization process, we remove them, making the cost function more receptive to errors in the calibration parameters. 



The periodicity-based formulation estimates the orientation and position of the rotation axis but leaves one degree-of-freedom unobservable along the axis direction. Specifically, a translation of the rotary frame-$P$ about its axis of rotation does not alter the periodicity cost defined in Eq.~(\ref{eq:periodicity_cost}). Consequently, the objective function is invariant to shifts along this direction, resulting in a one-dimensional family of equivalent solutions. In our parameterization, this invariance manifests as a coupled ambiguity between $t_x$ and $t_y$. To remove this gauge freedom and ensure a well-posed optimization problem, we fix $t_y$ to the initial calibration value, and treat $t_x$ as the remaining optimization variable. This reduces the number of free parameters from five to four.


For numerical optimization, we employ the Trust-Region Reflective algorithm from the SciPy library. The search space is constrained to $\pm 2$ cm for translation and $\pm 2^\circ$ for rotation about each degree of freedom. The termination tolerances are set to ${\texttt{xTol} = 10^{-12}, \ \texttt{fTol} = 10^{-10}}$, with a maximum of 10 function evaluations per inner optimization and up to 15 iterations in Algorithm~\ref{alg:periodicity_algo}.

\begin{figure}[t]
  \centering
  \begin{subfigure}[b]{0.15\textwidth}
    \centering
        \includegraphics[width=0.65\linewidth]{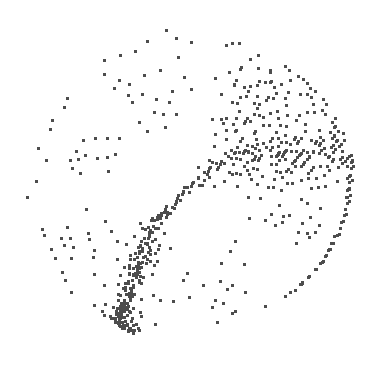}
    \caption{Iteration-1.}
  \end{subfigure}
  \hfill
  \begin{subfigure}[b]{0.15\textwidth}
    \centering
    \includegraphics[width=0.65\linewidth]{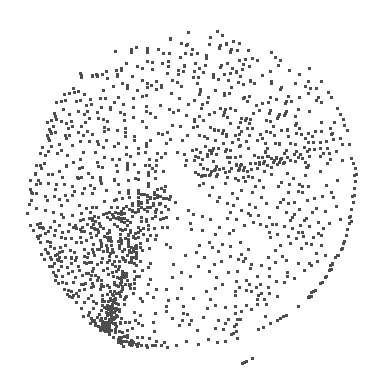}  
    \caption{Iteration-3.}
  \end{subfigure}
    \hfill
  \begin{subfigure}[b]{0.15\textwidth}
    \centering
    \includegraphics[width=0.65\linewidth]{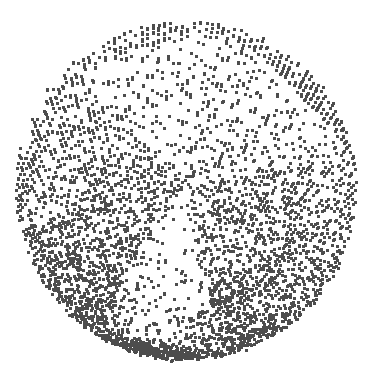}
    \caption{Iteration-10.}
  \end{subfigure}
  \caption{Number of matched points on the face of the target at different iterations of the periodicity-constrained fine-tuning algorithm.}
  \label{fig:periodicity_iterations}
  \vspace{-0.2in}
\end{figure}

Figure~\ref{fig:periodicity_iterations} illustrates the evolution of the number of matched points between $\mathcal{X}$ and $\mathcal{X}'$ across successive iterations on the target. The monotonic increase in correspondences indicates progressive improvement in the calibration estimate.

%% file: contents/experiments.tex
\section{Experiments and Results}
\label{sec:results}
This section presents a qualitative and quantitative overview of the experiments conducted for assessing the validity of the calibration algorithms. 


\subsection{Hardware Setup}
\label{subsec:mechanical}
\begin{figure}[t]
  \centering
  \includegraphics[width=0.4\linewidth, trim={0 0 0 0cm},clip]{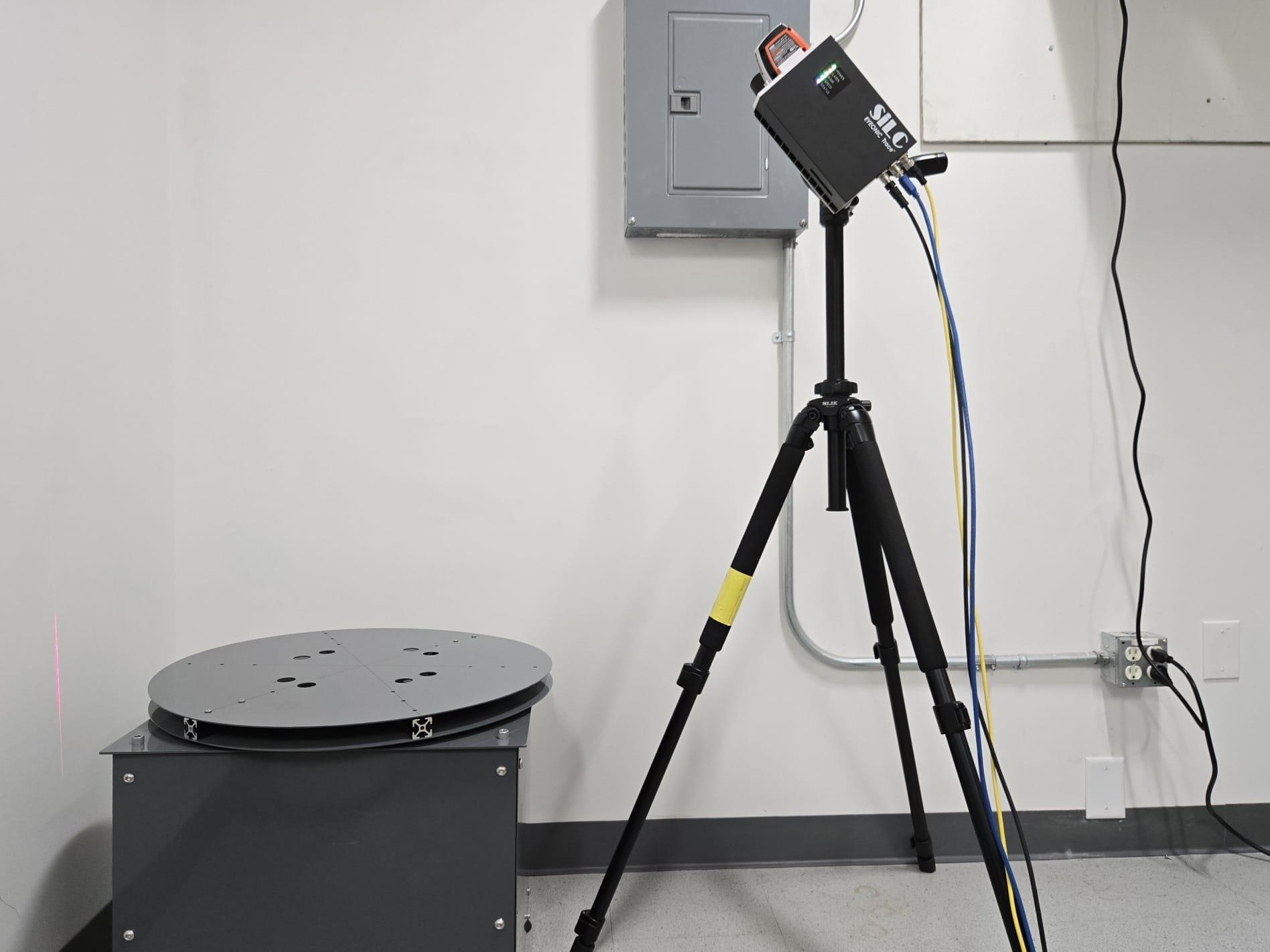}  
  \caption{Rotary platform and tripod mounted lidar setup.}
  \label{fig:mech_setup}
\end{figure}

We built a turntable setup with a 58.5 cm diameter circular metallic plate being rotated by a Clearpath-MCVC DC servo motor, with a maximum speed rating of 1130 rpm and peak torque of 9.9 N-m. To accurately capture the rotation, a Lika Electronic incremental (quadrature) rotary encoder is coaxially mounted with a resolution of 1024 ppr. The object on the turntable is scanned by a tripod-mounted Eyeonic Trace laser line scanner, which is an FMCW lidar from SiLC\footnote{\scriptsize\url{https://www.silc.com/eyeonic-trace/}}. It is rated at 1 mm precision and has $72\degree$ FOV, delivering about 1 million points per second. The full mechanical setup is shown in Fig.~\ref{fig:mech_setup}.

\subsection{Data}
\label{subsec:data}

We collected nine datasets from the rotary platform at varying lidar
pitch orientations and ranges, summarized in
Table~\ref{tab:datasets}. These configurations reflect practical
object-scanning setups, where the lidar is mounted in close proximity
to obtain good resolution on the object of interest while respecting
the minimum-range specification of the device.

\begin{table}[t]
    \centering
    \caption{Approximate range and pitch angle for datasets.}
    \begin{tabular}{l|ccc}
        \toprule
         Range and \\ Pitch angle & 1.00 m & 1.25 m & 1.50 m \\ 
        \midrule
        30\degree & $\mathcal{D}_{1}$ & $\mathcal{D}_{2}$ & $\mathcal{D}_{3}$ \\
        45\degree & $\mathcal{D}_{4}$ & $\mathcal{D}_{5}$ & $\mathcal{D}_{6}$ \\
        60\degree & $\mathcal{D}_{7}$ & $\mathcal{D}_{8}$ & $\mathcal{D}_{9}$ \\
        \bottomrule
    \end{tabular}
    \label{tab:datasets}
\end{table}

\subsection{Metrics}
\label{subsec:metrics}

In practical settings, the true calibration parameters are not directly observable without specialized, expensive equipment, so we instead rely on indirect measures of quality. Our first metric is the dimensioning accuracy of a standard 10 cm cube. Because calibration errors manifest as geometric distortions in the point cloud, the deviation between the estimated and true side length of the cube provides a meaningful measure of calibration accuracy. To faithfully capture these distortions, we deliberately avoid any smoothing of the point cloud prior to estimating the cube dimensions.


The output of the fine‑tuning algorithms depends on the initial calibration provided to them. We therefore evaluate their precision across a range of initial conditions. Higher precision i.e., smaller variation in the resulting calibration for different initializations, indicates reduced sensitivity to initialization and is a desirable property of the fine‑tuning.

\subsection{Results}
\label{subsec:quantitative}
\subsubsection{Initial Calibration}
We quantify the \textit{error} in estimating edge length of a 10 cm cube, i.e. $e = \max_{i}\big(\hat{s}_i - s\big)$, where $s$ denotes the true edge length, and $\hat{s}_i$ denotes estimated length of $i^{\text{th}}$ edge. Table~\ref{tab:median_cube_dim_error_no_sample} denotes the median \textit{error} obtained during initial calibration across the nine datasets. 
The positive median error reflects the fact that we use the unsmoothed point cloud to construct an enclosing bounding box, which biases the estimate toward a larger box. Applying a smoothing method such as a moving average or Gaussian kernel smoother could substantially reduce this error, but that is not the focus of this work. Using the initial calibration alone, we obtain a median percentage \textit{error} of 5.1\%.
\begin{table}[t]
    \centering
    \caption{Median \textit{error} in estimating dimension of 10 cm cube using various calibration algorithms across the nine datasets from Table~\ref{tab:datasets}. Best value are denoted in \textbf{bold}.}
    \begin{tabular}{l|c}
        \toprule
         & Edge estimate \textit{error} (mm)\\
        \midrule
        Initial calibration & 5.1 \\
        Shape-constrained fine-tuning & 3.2 \\
        Periodicity-constrained fine-tuning & \textbf{2.7} \\
        \bottomrule
    \end{tabular}
    \label{tab:median_cube_dim_error_no_sample}
\end{table}

\subsubsection{Fine-tuning Calibration}
For the first set of experiments, we initialize the fine-tuning algorithm with initial calibration results. The extrinsic parameters thus obtained are used to estimate the edge length of a 10 cm cube and the median \textit{error} is reported in Table~\ref{tab:median_cube_dim_error_no_sample}. We observe that both fine-tuning methods systematically reduce the median \textit{error}, however, the periodicity-constrained fine-tuning outshines the shape-constrained fine-tuning, yielding a median percentage \textit{error} of 2.7\%.


In the second set of experiments, we evaluate the robustness of the fine‑tuning algorithm to initialization. To this end, we define two perturbation sets: set-$\mathbf{A}$, with translation limits of $\pm100$ mm and rotation limits of $\pm1^\circ$, and set-$\mathbf{B}$, with translation limits of $\pm200$ mm and rotation limits of $\pm2^\circ$. From each set, we sample 50 uniformly distributed perturbations and add them to the parameters obtained from the initial calibration, using the result as the initialization for the fine‑tuning algorithm. These bounds are chosen to reflect the anticipated error of the initial calibration and are consistent with our experience in practical data collection scenarios. Note that, during sampling, the value of $t_y$ is fixed.

\begin{figure}[t]
  \centering
  \begin{subfigure}{0.48\linewidth}
    \centering
    \includegraphics[width=0.8\linewidth]{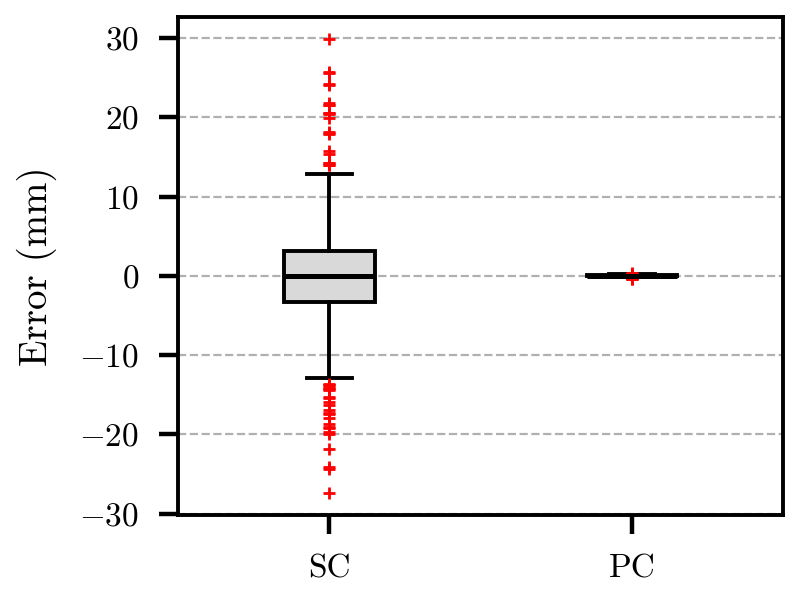}

    \subcaption{$t_x$ variation for set-$\mathbf{A}$.}
    \label{fig:translation_x_set_a}
  \end{subfigure}
  \hfill
  \begin{subfigure}{0.48\linewidth}
    \centering
    \includegraphics[width=0.8\linewidth]{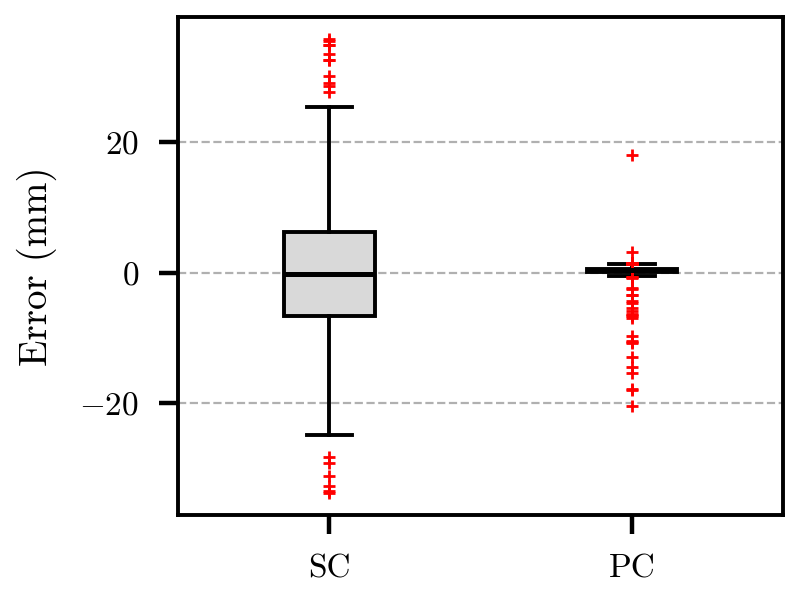}
    \subcaption{$t_x$ variation for set-$\mathbf{B}$.}
    \label{fig:translation_x_set_b}
  \end{subfigure}

  \begin{subfigure}{0.48\linewidth}
    \centering
    \includegraphics[width=0.8\linewidth]{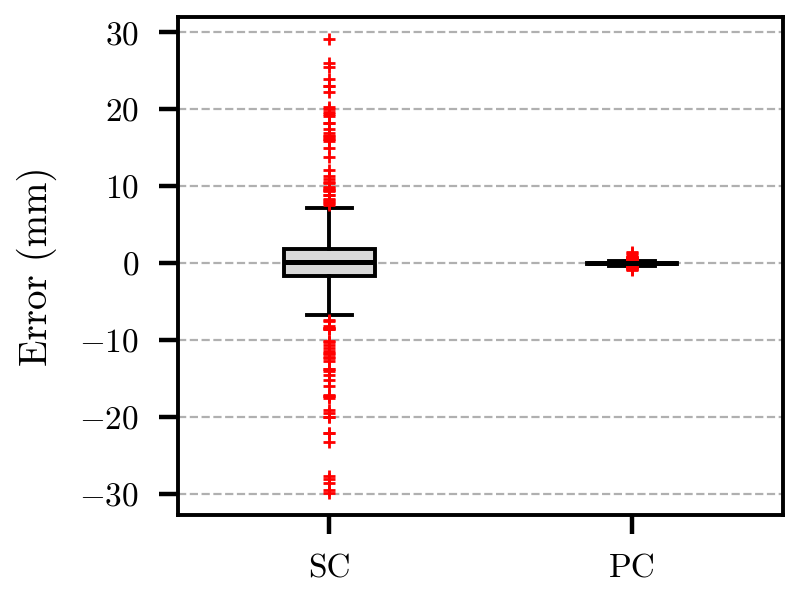}

    \subcaption{$t_z$ variation for set-$\mathbf{A}$.}
    \label{fig:translation_z_set_a}
  \end{subfigure}
  \hfill
  \begin{subfigure}{0.48\linewidth}
    \centering
    \includegraphics[width=0.8\linewidth]{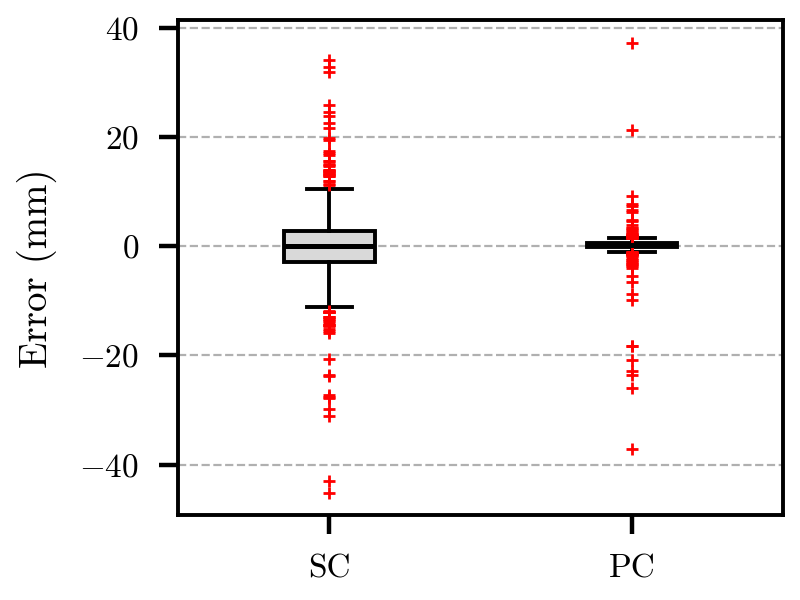}
    \subcaption{$t_z$ variation for set-$\mathbf{B}$.}
    \label{fig:translation_z_set_b}
  \end{subfigure}

    \begin{subfigure}{0.48\linewidth}
    \centering
    \includegraphics[width=0.8\linewidth]{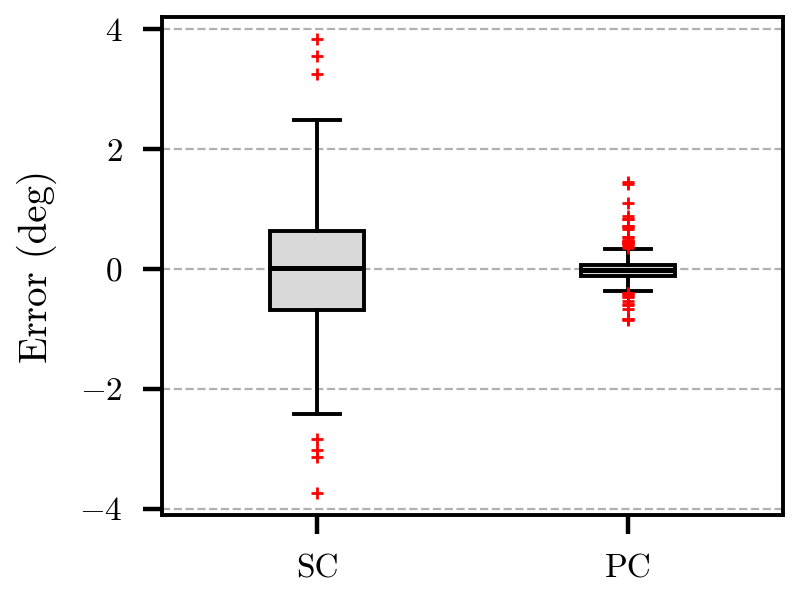}
    \subcaption{$r_y$ variation for set-$\mathbf{A}$.}
    \label{fig:rotational_y_set_a}
  \end{subfigure}
  \hfill
  \begin{subfigure}{0.48\linewidth}
    \centering
    \includegraphics[width=0.8\linewidth]{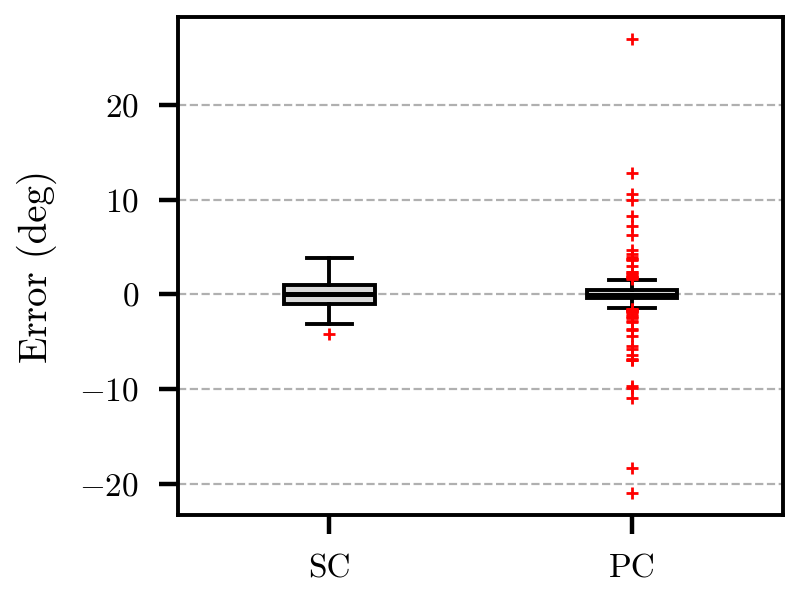}
    \subcaption{$r_y$ variation for set-$\mathbf{B}$.}
    \label{fig:rotational_y_set_b}
  \end{subfigure}

  \begin{subfigure}{0.48\linewidth}
    \centering
    \includegraphics[width=0.8\linewidth]{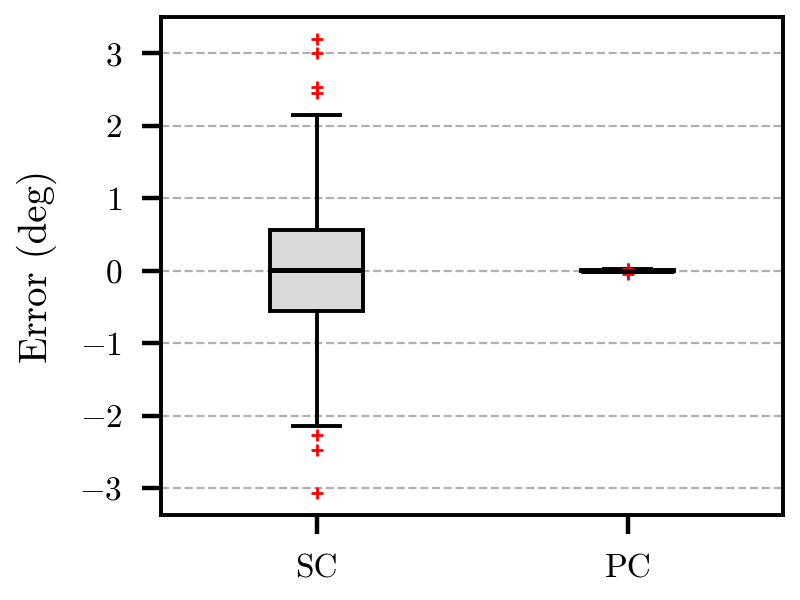}
    \subcaption{$r_z$ variation for set-$\mathbf{A}$.}
    \label{fig:rotational_z_set_a}
  \end{subfigure}
  \hfill
  \begin{subfigure}{0.48\linewidth}
    \centering
    \includegraphics[width=0.8\linewidth]{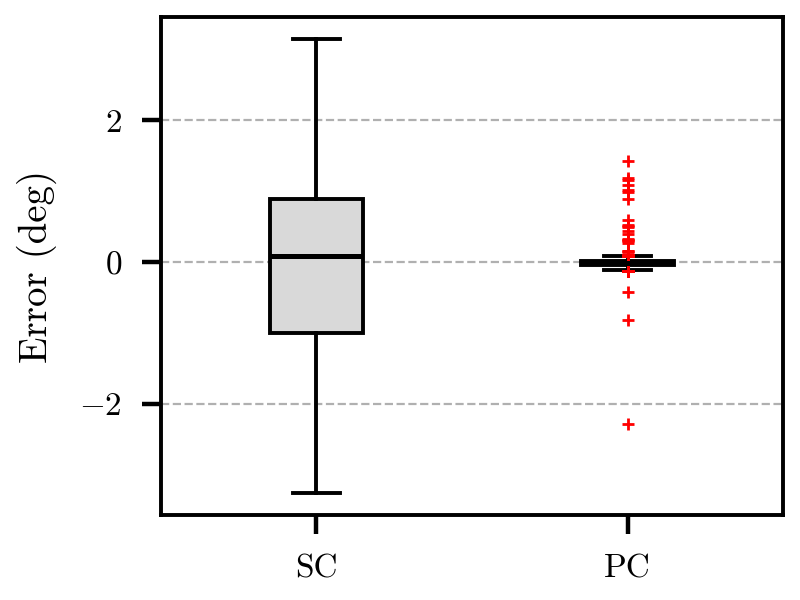}
    \subcaption{$r_z$ variation for set-$\mathbf{B}$.}
    \label{fig:rotational_z_set_b}
  \end{subfigure}
  
  \caption{Deviation of translational and rotational values from their \textit{mean} estimate, obtained by feeding initial calibration results to the fine-tuning algorithms, with 50 sampled perturbations from set-$\mathbf{A}$ and set-$\mathbf{B}$ ($N=450$ points).}
  \label{fig:translational_rotational_box_plots}
\end{figure}

Fig.~\ref{fig:translation_x_set_a} and \ref{fig:translation_x_set_b} show the deviation of converged value of $t_x$ from the \textit{mean} for the two fine-tuning algorithms for translation in $x$-axis. Under the sampled initial conditions from set-$\mathbf{A}$, the periodicity-constrained algorithm achieves sub-millimeter variation, whereas the shape-constrained approach can exhibit deviation up to 12 mm, excluding the outliers. 
When the range of initial perturbation is doubled (set-$\mathbf{B}$), the periodicity-constrained algorithm maintains variation in $t_x$ upto 1 mm for the majority of trials, with only a few outliers exceeding this threshold.
A similar trend is observed for the $z$-axis, as shown in Fig.~\ref{fig:translation_z_set_a} and \ref{fig:translation_z_set_b}. 

Fig~\ref{fig:rotational_y_set_a}, \ref{fig:rotational_y_set_b}, \ref{fig:rotational_z_set_a} and \ref{fig:rotational_z_set_b}, show a comparison for rotational deviation along $y$-axis and $z$-axis, respectively. The shape-constrained algorithm exhibits higher dispersion, as indicated by a much larger interquartile range and longer whiskers. In contrast, the periodicity-constrained algorithm shows a more tightly concentrated central distribution. 
Especially, if the initial error is within the limits of set-$\mathbf{A}$, then the outliers are significantly reduced and are of smaller magnitude. 

\begin{figure}[t]
  \centering
  \begin{subfigure}{0.48\linewidth}
    \centering
    \includegraphics[width=0.8\linewidth]{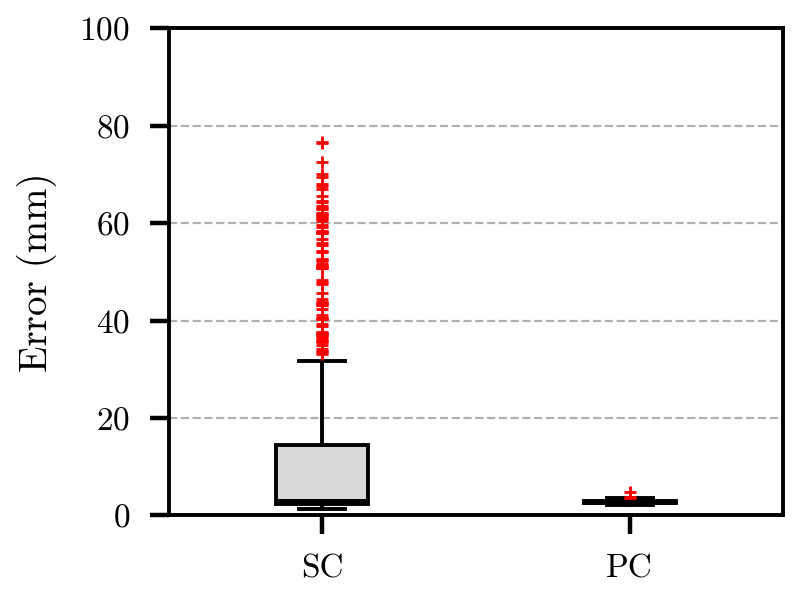}
    \subcaption{Cube edge \textit{error} for set-$\mathbf{A}$.}
    \label{fig:edge_1_set_a}
  \end{subfigure}
  \hfill
  \begin{subfigure}{0.48\linewidth}
    \centering
    \includegraphics[width=0.8\linewidth]{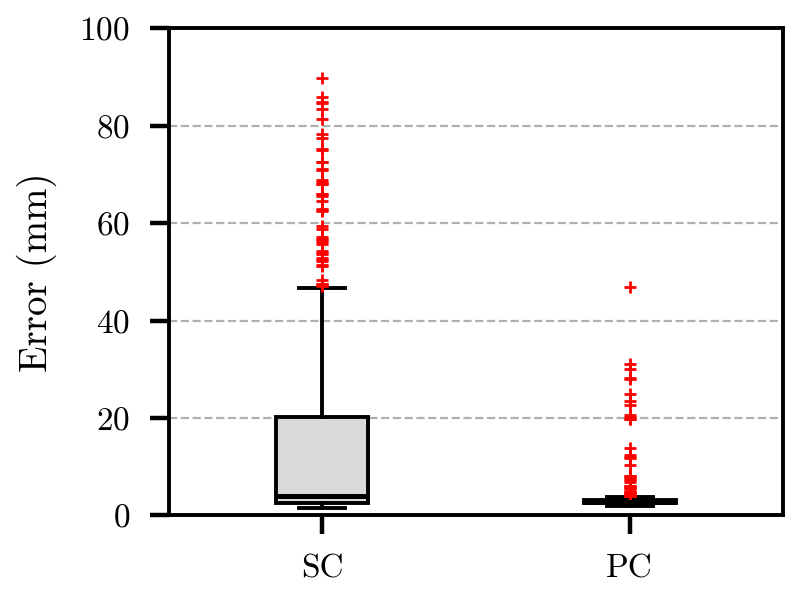}
    \subcaption{Cube edge \textit{error} for set-$\mathbf{B}$.}
    \label{fig:edge_1_set_b}
  \end{subfigure}
  \caption{Cube dimensioning \textit{error} obtained by feeding initial calibration results to the fine-tuning algorithms, with 50 sampled perturbations from set-$\mathbf{A}$ and set-$\mathbf{B}$ ($N=450$ points).}
  \label{fig:edge_1}
  \vspace{-0.1in}
\end{figure}

Fig.~\ref{fig:edge_1} illustrates the \textit{error} in estimating the edge length of a 10 cm cube. Table~\ref{tab:median_cube_dim_error} reports the median error for the different calibration methods. Compared to using the initial calibration alone, both fine‑tuning algorithms reduce the median error, with the periodicity‑constrained method achieving roughly a twofold improvement. Moreover, for initial perturbations sampled from set-$\mathbf{A}$ and set-$\mathbf{B}$, the periodicity‑constrained algorithm exhibits an order‑of‑magnitude lower variation in the estimates than the shape‑constrained approach, as shown in Fig.~\ref{fig:edge_1}, showing a higher tolerance to initial calibration errors.



\begin{table}[t]
    \centering
    \caption{Median \textit{error} in estimating 10 cm cube dimensions using calibration algorithms across nine datasets from Table~\ref{tab:datasets} and sampled initial conditions. Best values are denoted in \textbf{bold}.}
    \begin{tabular}{l|cc}
        \toprule
         & \multicolumn{2}{c}{Edge estimate Error (mm)}  \\ 
         & set-\textbf{A} & set-\textbf{B} \\
        \midrule
        Initial calibration & 5.1 & 5.1 \\
        Shape-constrained fine-tuning & 3.0 & 4.0 \\
        Periodicity-constrained fine-tuning & \textbf{2.7} & \textbf{2.8} \\
        \bottomrule
    \end{tabular}
    \label{tab:median_cube_dim_error}
    \vspace{-0.1in}
\end{table}

\begin{figure}[!t]
  \centering
  \begin{subfigure}{0.48\linewidth}
    \centering
    \includegraphics[width=0.8\linewidth]{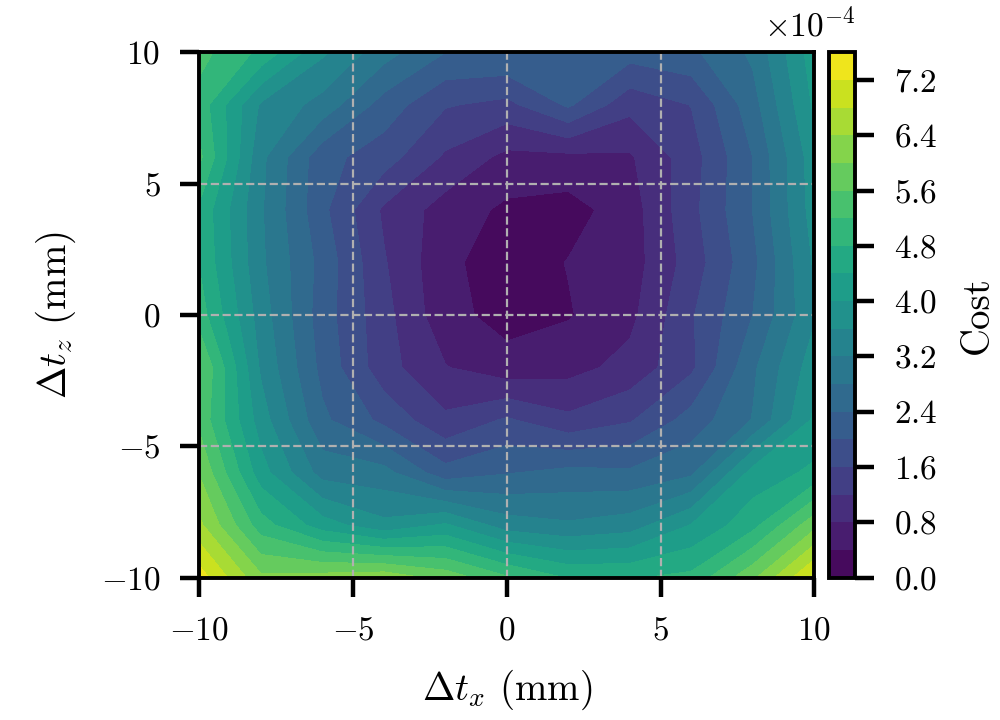}

    \subcaption{SC cost for translation.}
    \label{fig:cost_sc_trans}
  \end{subfigure}
  \hfill
  \begin{subfigure}{0.48\linewidth}
    \centering
    \includegraphics[width=0.8\linewidth]{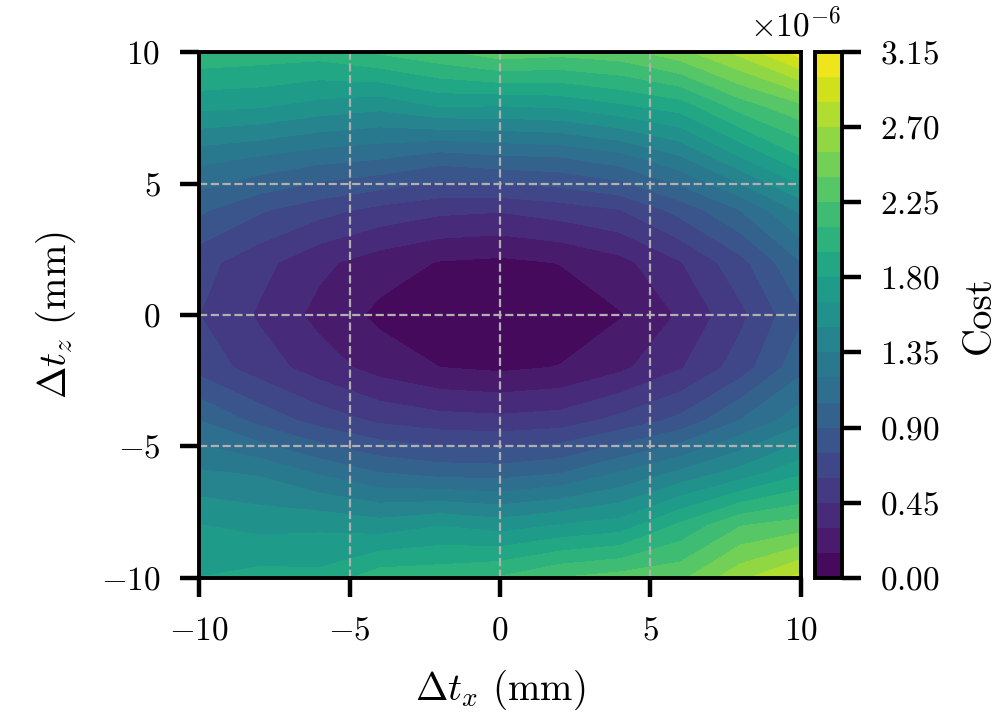}
    \subcaption{PC cost for translation.}
    \label{fig:cost_pc_trans}
  \end{subfigure}

  \begin{subfigure}{0.48\linewidth}
    \centering
    \includegraphics[width=0.8\linewidth]{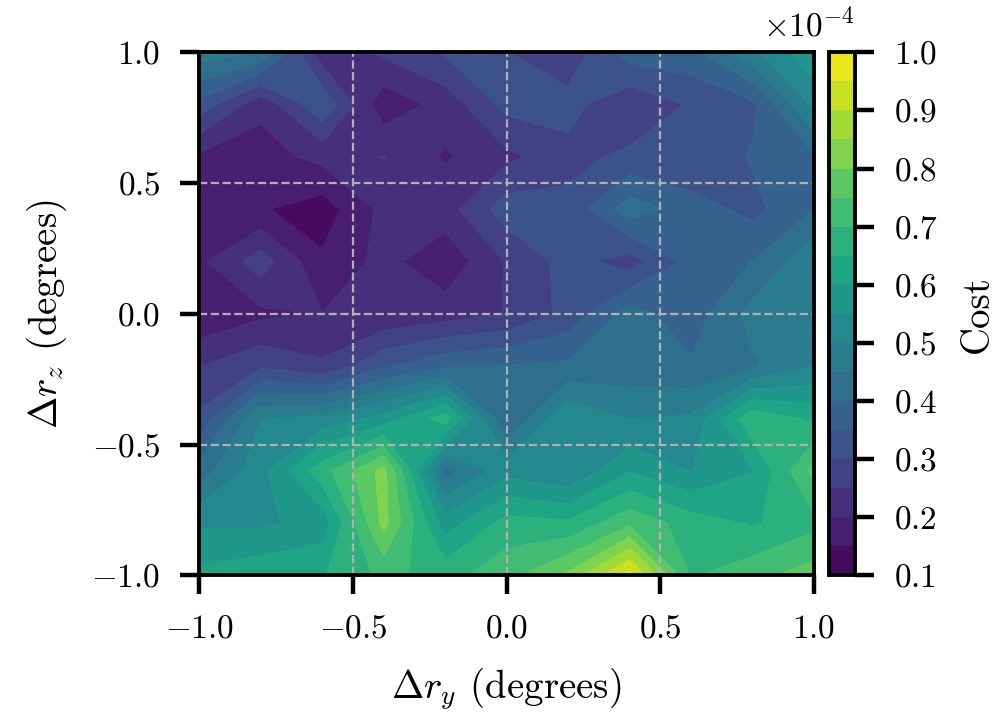}

    \subcaption{SC cost for rotation.}
    \label{fig:cost_sc_rot}
  \end{subfigure}
  \hfill
  \begin{subfigure}{0.48\linewidth}
    \centering
    \includegraphics[width=0.8\linewidth]{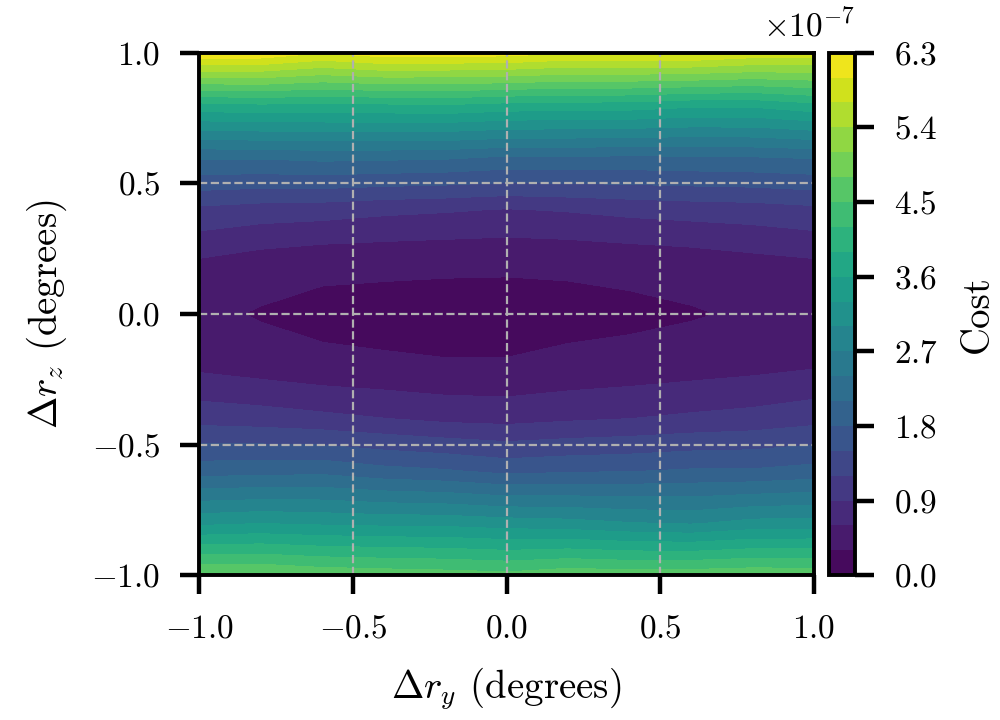}
    \subcaption{PC cost for rotation.}
    \label{fig:cost_pc_rot}
  \end{subfigure}  
  \caption{Cost landscape for fine-tuning obtained by varying translation and rotation values for a randomly selected dataset $D_6$.}
  \label{fig:cost_sc_pc}
\end{figure}
\subsection{Discussion}
\label{subsec:discussion}
The differences in convergence characteristics of the two fine-tuning algorithms motivated a closer investigation into the cost functions used in the two approaches. Based on a randomly selected dataset $D_6$ from our set of collected sequences, we show the cost value as a function of the translation and the rotation parameters in Fig.~\ref{fig:cost_sc_pc}. The axes represent deviation from the converged value in the respective fine-tuning approaches.

The translation cost landscape for both algorithms appears relatively smooth, with a clearly identifiable minimum in the local vicinity. Still, the elliptical shape of the contour lines in the periodicity-constrained fine-tuning, shown in Fig.~\ref{fig:cost_pc_trans}, is more desirable for a nonlinear optimizer, compared to the lines in the shape-constrained fine-tuning, shown in Fig.~\ref{fig:cost_sc_trans}. The differences, however, in the rotational cost landscape are pronounced in Fig~\ref{fig:cost_sc_rot} and ~\ref{fig:cost_pc_rot}. The cost for the shape-constrained algorithm is noticeably more irregular, with local undulations which can cause the optimizer to become trapped in a local minimum. This explains the significant dispersion in rotational values observed in Fig.~\ref{fig:rotational_y_set_a}, \ref{fig:rotational_y_set_b}, \ref{fig:rotational_z_set_a}, and  \ref{fig:rotational_z_set_b}. In contrast, the periodicity-constrained algorithm's cost function has gentler gradients and a clear minimum. We observed that the cost for periodicity-constrained algorithm is more sensitive to $r_z$ than to $r_y$, which explains the smaller spread in the post fine-tuning $r_z$ values in Fig.~\ref{fig:rotational_z_set_a}, and \ref{fig:rotational_z_set_b}.

\begin{figure}[t]
  \centering
  \begin{subfigure}[b]{0.25\linewidth}
    \centering
    \includegraphics[width=0.8\linewidth,clip]{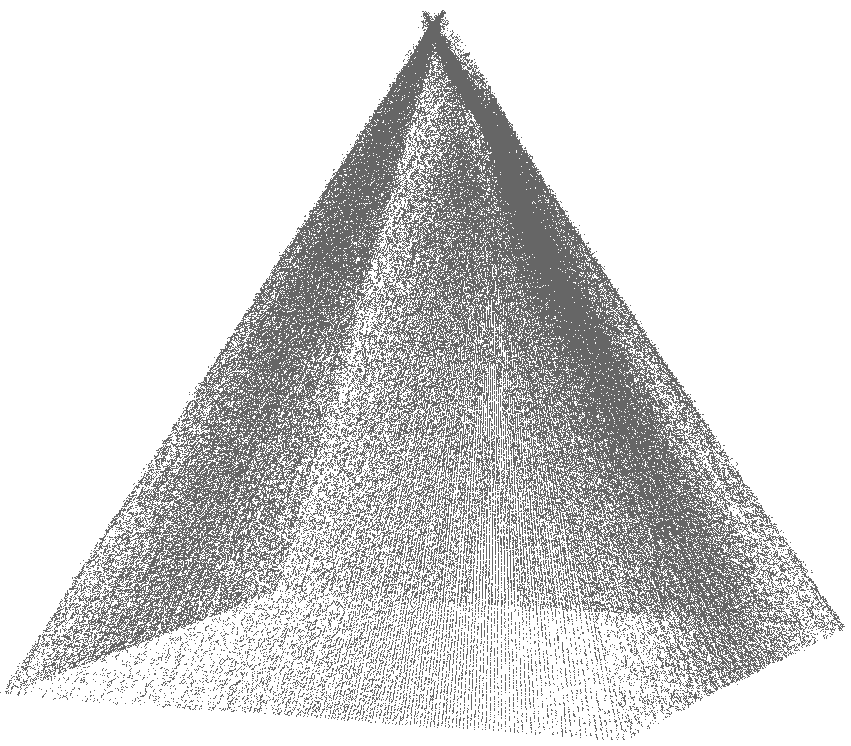}
    \caption{IC.}
  \end{subfigure}
  \hfill
  \begin{subfigure}[b]{0.25\linewidth}
    \centering
    \includegraphics[width=0.8\linewidth,clip]{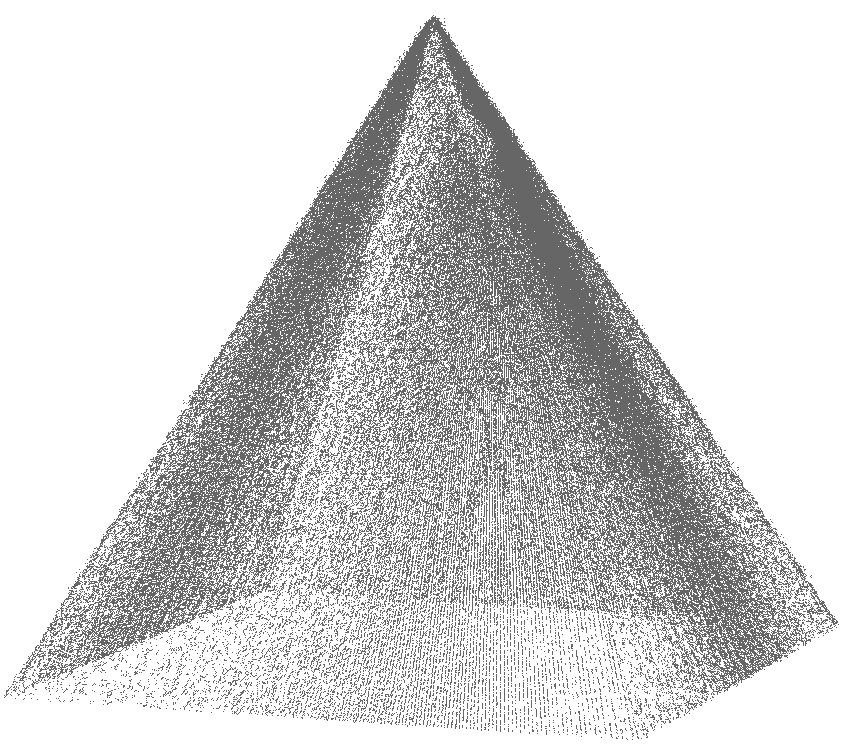}
    \caption{PC.}
  \end{subfigure}
    \hfill
  \begin{subfigure}[b]{0.22\linewidth}
    \centering
    \includegraphics[width=\linewidth,trim={10cm 20cm 10cm 0cm},clip]{images/qualitative_pyramid_initial.png}
    \caption{Zoomed IC.}
  \end{subfigure}
  \begin{subfigure}[b]{0.22\linewidth}
    \centering
    \includegraphics[width=\linewidth,trim={10cm 20cm 10cm 0cm},clip]{images/qualitative_pyramid_finetune.png}
    \caption{Zoomed PC.}
  \end{subfigure}
  \caption{Pyramid (15 cm$\times$15 cm$\times$15 cm) visualization with initial calibration (IC) and periodicity-constrained (PC) fine-tuning.}
  \label{fig:qualitative_pyramid}
  \vspace{-0.1in}
\end{figure}

\begin{figure}[t]
  \centering
  \begin{subfigure}[b]{0.25\linewidth}
    \centering
    \includegraphics[width=0.8\linewidth,clip]{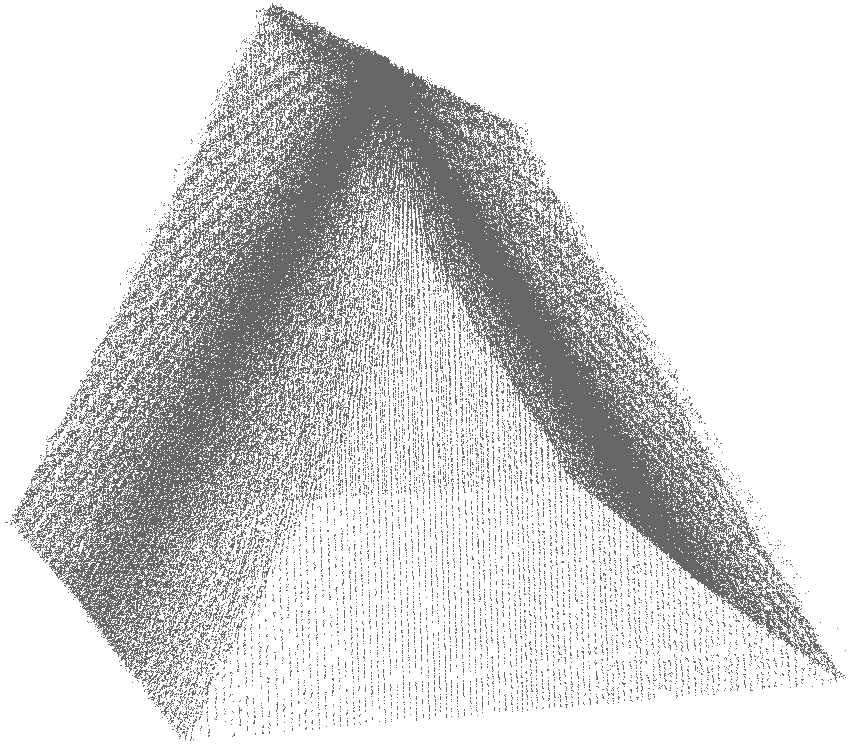}
    \caption{IC.}
  \end{subfigure}
  \hfill
  \begin{subfigure}[b]{0.25\linewidth}
    \centering
    \includegraphics[width=0.8\linewidth,clip]{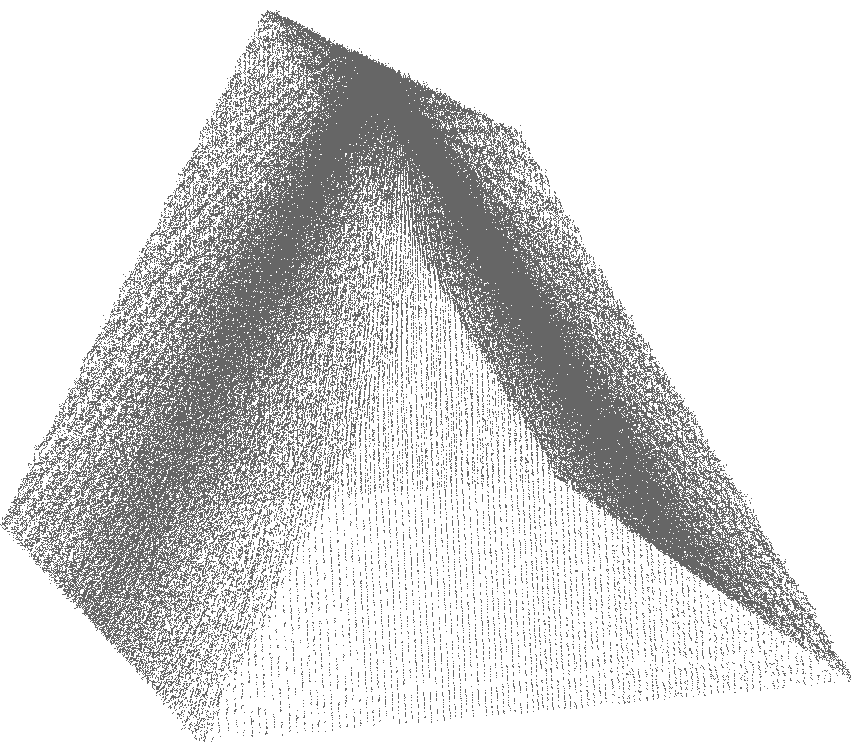}
    \caption{PC.}
  \end{subfigure}
    \hfill
  \begin{subfigure}[b]{0.22\linewidth}
    \centering
    \includegraphics[width=\linewidth,trim={18cm 1cm 0cm 16cm},clip]{images/qualitative_prism_initial.png}
    \caption{Zoomed IC.}
  \end{subfigure}
  \begin{subfigure}[b]{0.22\linewidth}
    \centering
    \includegraphics[width=\linewidth,trim={18cm 1cm 0cm 16cm},clip]{images/qualitative_prism_finetune.png}
    \caption{Zoomed PC.}
  \end{subfigure}
  \caption{Prism (15 cm edge $\times$ 22.5 cm depth) visualization with initial calibration (IC) and periodicity-constrained (PC) fine-tuning.}
  \label{fig:qualitative_prism}
  \vspace{-0.1in}
\end{figure}

\begin{figure}[!t]
  \centering
  \begin{subfigure}[b]{0.25\linewidth}
    \centering
    \includegraphics[width=0.8\linewidth,clip]{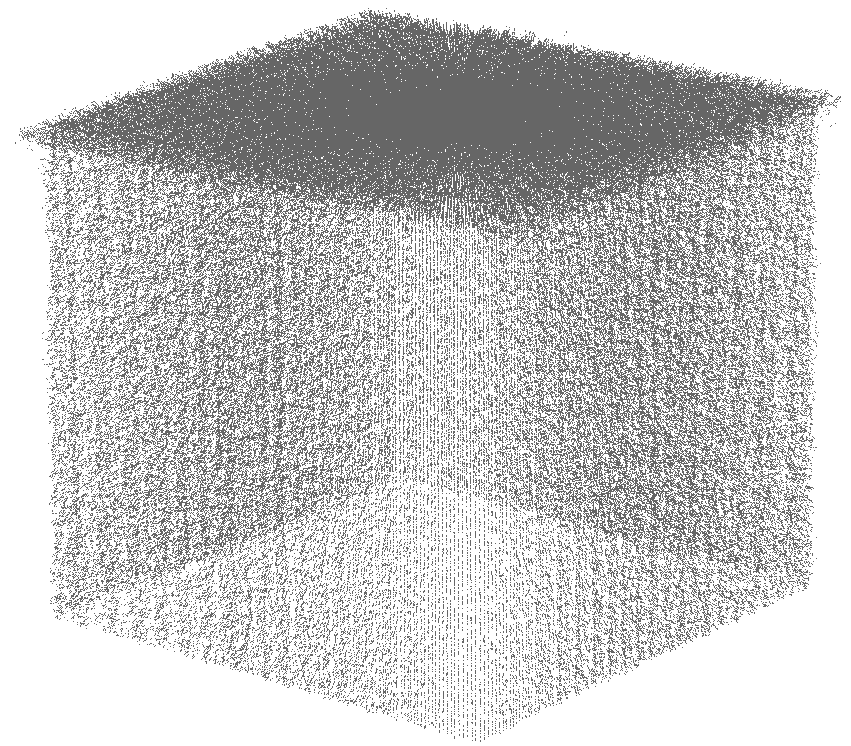}
    \caption{IC.}
  \end{subfigure}
  \hfill
  \begin{subfigure}[b]{0.25\linewidth}
    \centering
    \includegraphics[width=0.8\linewidth,clip]{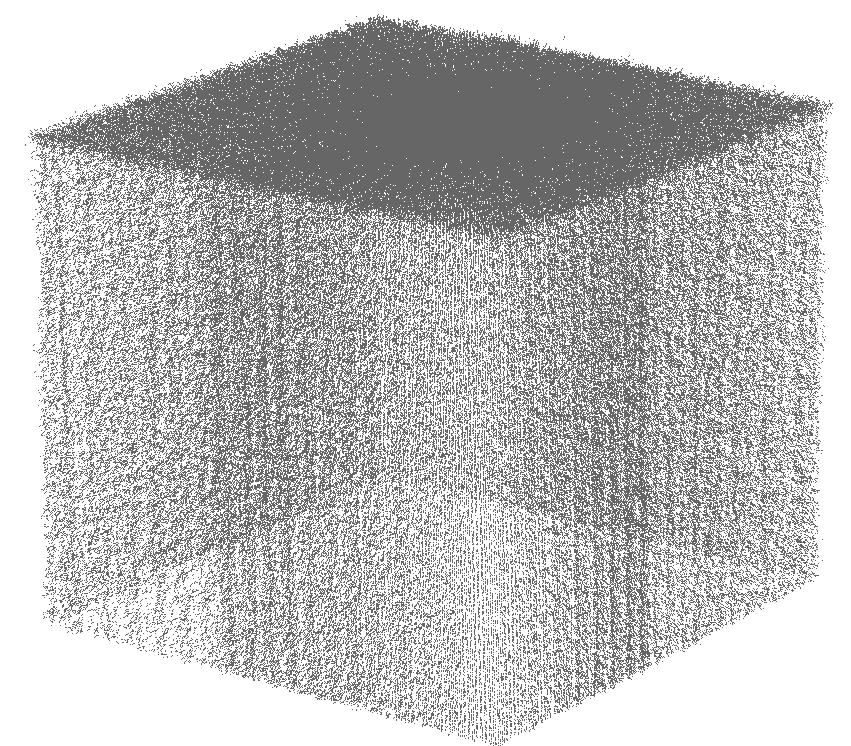}
    \caption{PC.}
  \end{subfigure}
    \hfill
  \begin{subfigure}[b]{0.22\linewidth}
    \centering
    \includegraphics[width=\linewidth,trim={22cm 18cm 0cm 2cm},clip]{images/qualitative_cube_initial.png}
    \caption{Zoomed IC.}
  \end{subfigure}
  \begin{subfigure}[b]{0.22\linewidth}
    \centering
    \includegraphics[width=\linewidth,trim={22cm 18cm 0cm 2cm},clip]{images/qualitative_cube_finetune.png}
    \caption{Zoomed PC.}
  \end{subfigure}
  \caption{Cube (10 cm) visualization with initial calibration (IC) and periodicity-constrained (PC) fine-tuning.}
  \label{fig:qualitative_cube}
  \vspace{-0.1in}
\end{figure}

To qualitatively compare the initial calibration with the fine‑tuned result, we scanned several objects, and their point clouds are shown in Figs.~\ref{fig:qualitative_pyramid},~\ref{fig:qualitative_prism},~and~\ref{fig:qualitative_cube}. For the pyramid in Fig.~\ref{fig:qualitative_pyramid}, the initially calibrated point cloud exhibits multiple peaks, which are removed after fine‑tuning. In Fig.~\ref{fig:qualitative_prism}, the initial calibration leads to misaligned points on one of the faces of the prism, whereas the fine‑tuned point cloud appears smooth. Finally, in Fig.~\ref{fig:qualitative_cube}, we observe corner bleeding in the initially calibrated point cloud that is substantially reduced after fine‑tuning. Together, these qualitative results highlight the benefits of the proposed two‑stage calibration.

The cube dimensioning results reported in Table~\ref{tab:median_cube_dim_error} are conservative, as they are based on a naive minimum-volume oriented bounding box fitted directly to the raw point cloud. In practical dimensioning applications, smoothing is applied prior to polyhedral fitting to reduce measurement variance. Applying a moving least squares filter to the cube point cloud reduced the median dimensioning error to 0.3 mm (0.3\%) for the nine datasets, representing an order-of-magnitude improvement over the calibrated but unsmoothed data.

In terms of computation, the initial calibration algorithm requires approximately 10 seconds. On the other hand, both shape-constrained and periodicity-constrained fine-tuning require up to 2 minutes to converge to the calibration parameters on an Intel i7 16-core (2.5 GHz) system. 

%% file: contents/conclusion.tex
\section{Conclusion}
\label{sec:conclusion}

In this work, we presented a two-stage calibration framework for estimating the extrinsic transformation between a line-scanning lidar and a rotating platform. The proposed approach combines a geometry-driven static initialization procedure with dynamic fine-tuning strategies that leverage structured target observations. The static stage provides a computationally efficient means of obtaining a reliable initial estimate, while the dynamic stage refines the parameters through constrained nonlinear optimization. Although the static initialization requires limited human intervention, the subsequent dynamic refinement stage compensates for these approximations through automated nonlinear optimization.

We investigated two fine-tuning strategies: shape-constrained and periodicity-constrained formulations both designed to exploit geometric structure in the acquired data. Experimental evaluations under varying initial conditions demonstrate that the periodicity-constrained approach exhibits robustness to significant initial calibration errors. This behavior is further supported by analysis of its cost landscape, which reveals a comparatively well-defined basin of convergence. These findings highlight the effectiveness of periodicity-driven constraints for rotary-platform calibration and suggest that dynamic redundancy can be systematically exploited to improve estimation stability.

Future work will focus on characterizing the basin of convergence of the periodicity-based formulation and investigating whether the static initial calibration stage can be eliminated altogether. Additional research directions include reformulating the shape-constrained objective to improve smoothness and optimizer performance, potentially through an outer-loop iterative scheme, and studying the influence of platform rotation speed on calibration accuracy.